\documentclass{article} %
\usepackage{iclr2027_conference,times}
\usepackage{amsmath,amssymb,amsthm}
\usepackage{mathtools}
\usepackage{graphicx}
\usepackage{xcolor}
\usepackage{tcolorbox}
\usepackage[colorlinks=true,linkcolor=blue,citecolor=blue,urlcolor=blue]{hyperref}
\usepackage{algorithm, algorithmic}

\newtheorem{theorem}{Theorem}[section]
\newtheorem*{theorem*}{Theorem}

\newtheorem*{proposition*}{Proposition}
\newtheorem{lemma}[theorem]{Lemma}
\newtheorem*{lemma*}{Lemma}

\newtheorem*{conjecture*}{Conjecture}

\newtheorem*{fact*}{Fact}

\newtheorem*{hypothesis*}{Hypothesis}

\newtheorem*{claim*}{Claim}

\theoremstyle{definition}
\newtheorem{definition}[theorem]{Definition}

\theoremstyle{remark}

\newtheorem*{remark*}{Remark}

\newcommand{\eat}[1]{}

\newcommand{\calX}{\mathcal{X}}

\newcommand{\norm}[1]{\lVert #1 \rVert}

\newcommand{\Esymb}{\mathbb{E}}
\newcommand{\Psymb}{\mathbb{P}}

\newcommand{\Exp}{\Esymb}
 
 \DeclareMathOperator*{\ProbOp}{\Psymb}

 \DeclareMathOperator*{\argmin}{argmin}

\renewcommand{\Pr}{\ProbOp}

\renewcommand{\epsilon}{\varepsilon}

\newcommand{\ReWD}{\RTW}
\newcommand{\diff}{\operatorname{d}}

\newcommand{\R}{\mathbb{R}}

\newcommand{\RTW}{\mathrm{RTW}}

\newif\ifnotes\notesfalse

\ifnotes
\usepackage{color}
\definecolor{mygrey}{gray}{0.50}
\newcommand{\notename}[2]{{\textcolor{purple}{\footnotesize{\bf (#1:} {#2}{\bf ) }}}}

\newcommand{\enote}[1]{{\notename{Bob}{#1}}}
\newcommand{\vnote}[1]{{\notename{Vaidehi}{#1}}}

\newcommand{\anote}[1]{{\notename{Aravindan}{#1}}}
\newcommand{\tnote}[1]{{\notename{Tanmay}{#1}}}
\newcommand{\dnote}[1]{{\notename{Dionysis}{#1}}}
\else

\newcommand{\notename}[2]{{}}

\newcommand{\enote}[1]{}
\newcommand{\vnote}[1]{}

\newcommand{\anote}[1]{}
\newcommand{\tnote}[1]{}
\newcommand{\dnote}[1]{}

\fi

\newcommand{\targ}{t}

\usepackage{xcolor}
\definecolor{biascolor}{RGB}{31,119,180}
\definecolor{regcolor}{RGB}{200,100,0}

\providecommand{\inparen}[1]{\left(#1\right)}

\title{Regularized Transport Weights: a Principled Approach for Active Domain Adaptation}

\iclrfinalcopy

\author{%
Anxin Guo$^{1,2,*}$\quad
Tanmay Sinha$^{1,2,*}$\quad
Dionysis Arvanitakis$^{1,2}$\quad
Vaidehi Srinivas$^{3}$\\
\bfseries Alex Drlica-Wagner$^{2,4,5,6}$\quad
Alexander P.~Ji$^{2,5,6}$\quad
Aravindan Vijayaraghavan$^{1,2}$%
}

\newcommand{\affilfootnote}{%
  {\let\thefootnote\relax
   \footnotetext{%
   $^{*}$Equal contribution.\\
   $^{1}$Department of Computer Science, Northwestern University, Evanston, IL, USA.\\
   $^{2}$NSF-Simons AI Institute for the Sky (SkAI), 172 E.~Chestnut St., Chicago, IL 60611, USA.\\
   $^{3}$Department of Electrical Engineering and Computer Science, MIT, Cambridge, MA, USA.\\
   $^{4}$Fermi National Accelerator Laboratory, Batavia, IL, USA.\\
   $^{5}$Department of Astronomy and Astrophysics, University of Chicago, Chicago, IL, USA.\\
   $^{6}$Kavli Institute of Cosmological Physics, University of Chicago, Chicago, IL, USA.}}}

\begin{document}
\maketitle
\affilfootnote
\lhead{Preprint}

\begin{abstract}
    Distribution shift, where there is a mismatch between training and test distributions, is a fundamental challenge in modern machine learning. 
    In this context, we consider the setting of active domain adaptation, where the learner has access to a labeled source dataset $S$, and aims to classify an unlabeled target dataset $T$.  The learner is allowed to query the labels of a few points in $T$ to mitigate the distribution shift between $S$ and $T$, as in active learning.
    We propose a principled, task-agnostic method that jointly selects target queries and reweighs the source and queried examples. Our method tries to effectively use the existing source labels in order to balance coverage of the target distribution while mitigating the variance in the labels.
    We achieve this by minimizing a natural Regularized Transport Weighting (RTW) objective %
    between the augmented dataset and the target $T$.
    We provide theoretical guarantees for our method, including a generalization bound and an approximation guarantee for our algorithm. 
    Experiments on standard benchmarks show that our method is competitive with the state-of-the-art methods, and outperforms prior methods on some standard benchmarks-- with only a few samples, we are able to train a classifier that significantly   improves upon the accuracy achieved on the source dataset alone. %

\end{abstract}

\section{Introduction}

\newcommand{\calH}{\mathcal{H}}
\newcommand{\calP}{\mathcal{P}}
\newcommand{\calQ}{\mathcal{Q}}
\newcommand{\calY}{\mathcal{Y}}

A central challenge in machine learning is to transfer learned concepts to new unseen data and domains. A mismatch between the training and test distributions, such as covariate shift, can substantially reduce prediction accuracy.  For example, a model trained on medical images from one hospital may perform poorly at another~\citep{quinonero2008book, mansour2009domain, Peng_2019_ICCV,koh21awilds}. %
It may be expensive to get labeled data from the target domain, but we often have access to data from related domains.  This motivates \emph{domain adaptation}, where the learner uses data from a source distribution $\calP$, to learn predictors for the target distribution $\calQ$. 

Concretely, $\calP$ and $\calQ$ are distributions over \((x, y) \in \calX \times \calY\) where \(\calX\) is the \emph{feature} space and \(\calY\) is the \emph{label} space.
We use $\calP_x$ and $\calQ_x$ to denote the feature marginals of $\calP$ and $\calQ$ respectively.  
We will use $\calH$ to denote a class of hypothesis functions $h : \calX \rightarrow \calY$ of interest. 
We consider the {\em covariate shift} setting between \(\calP\) and \(\calQ\), where the marginal distributions $\calP_x$ and $\calQ_x$ differ, while the conditional distribution $Y|x$ is the same for both $\calP$ and $\calQ$. 

Domain adaptation is naturally limited by how different the source and target distributions are from one another; in regions of the covariate space where $\calP$ and $\calQ$ differ significantly, the source distribution $\calP$ alone may provide insufficient information to make good predictions. %
A popular approach to tackle covariate shift is importance sampling~\citep{sugiyama2007kliep, quinonero2008book}. Importance sampling uses unlabeled data from the target distribution $\calQ$  to reweigh source examples according to the target-to-source density ratio. Estimating importance weights can be difficult in high dimensions. Moreover even when estimated exactly, large importance weights can lead to high statistical error. 
These challenges motivate \emph{active domain adaptation} methods, where the learner can query the labels of a small number of target points, in addition to unlabeled data from $\calQ$ and labeled data from the source $\calP$. 

\paragraph{Active Domain Adaptation.}
Active domain adaptation captures a setting where target labels are expensive but not impossible to acquire. %
In the context of active domain adaptation, the question becomes: \begin{center}{\em Which target labels should we acquire and how should we combine them with the source data?}\end{center} The  ideal method should learn as much information about $\calQ$ as possible from $\calP$, and spend its query budget on parts of $\calQ$ where we have insufficient information from the source $\calP$.

We focus on {\em task-agnostic query methods}: for a fixed feature representation (and potentially a hypothesis class), queries are selected only using the features of the source and target datasets; label information is not used in determining the query points.  
This allows the selected points to be reused across multiple tasks using the same queries. %
This is motivated by scientific domains like astronomy, where the queried information is rich, expensive information (e.g., high-resolution spectra) that can be used to provide ``labels'' for several different tasks. Task-agnostic query methods also fall into the class of single-shot active domain adaptation methods that have been studied previously \citep[e.g.,][]{deMathelin2022discrepancy, mahmood2022lowbudget}.  

\begin{figure}[t]
    \centering
    \includegraphics[width=\linewidth]{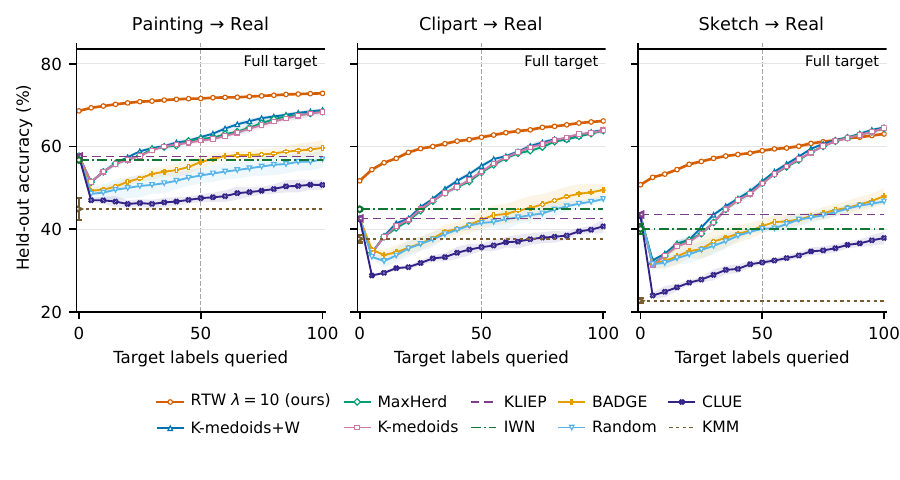}

    \caption{Held-out accuracy on the three mini-DomainNet tasks with Real as the target domain. RTW uses $\lambda=10$. Full target denotes the accuracy of a classifier trained on the entire target query pool, and horizontal lines show zero-query baselines that use no target labels. With $k=50$ queries, RTW improves accuracy by 14--20 percentage points over training on the source alone, and even at $k=0$ it outperforms other zero-query baselines through reweighting.}
    \label{fig:minidomainnet-three-to-real}
\vspace{-10pt}
\end{figure}

\subsection{Model and Our Contributions.} 
Formally, the learner is given an \emph{unlabeled} target dataset \(T \subset \calX \) of i.i.d. samples from \(\calQ_x\), a labeled source dataset $(s_1, y_1), (s_2, y_2) , \dots, (s_n,y_n) \in \calX \times \calY$ drawn from \(\calP\), and a query budget $k$. 
We use $S=\{s_1, \dots, s_n\} \subset \calX$ to denote the source dataset. 
A cost function $c: \calX \times \calX \to \mathbb{R}_{\ge 0}$ specifies the distance or cost between points in the feature space $\calX$ and encodes how much information $s$ provides about $t$. A natural choice is the Euclidean distance $c(s,t) = \| s-t\|_2$ over $\calX=\mathbb{R}^d$, or any metric over $\mathcal{X}$; in Section~\ref{sec:conditional-risk} we will use a choice that also depends on the hypothesis class $\calH$ to get better bounds.   

The learner is allowed to \emph{query} the labels of $k$ samples $A_k \subset T$. Moreover,  the learner can also pick the distribution i.e., weights $w=(w_s : s \in S \cup A_k)$ over the labeled samples corresponding to $S \cup A_k$. The classifier is then trained on the distribution of labeled examples corresponding to $S \cup A_k$ and $w$. %
Our goal is to find the best $k$ points to query along with the weights $w$ to optimize the loss on the target dataset. 
To this end, we introduce Regularized Transport Weighting (RTW), which jointly selects target queries $A_k$ and the weights on the source and queried examples $S \cup A_k$. 

\paragraph{Regularized Transport Weighting (RTW).} Our first theoretical contribution is to frame the problem of choosing queries from $T$ as minimizing a natural \emph{transport} objective. 
The objective trades off how well the weighted data represent the target against the statistical error from concentrating weight on too few samples.

\begin{tcolorbox}
\begin{definition}
Given source points $S$, target points $T$, select $k$ points $A_k \subseteq T$ and weights $w=(w_s: s \in S \cup A_k)$ that  
\begin{equation}
    \min_{\substack{A_k \subseteq T, |A_k| \le k}}
    \RTW_\lambda(S\cup A_k),
    \label{eq:intro-query-selection}
\end{equation}
where $\RTW_\lambda$ is the following regularized transport objective that jointly optimizes the weights $w=(w_s: s \in S \cup A_k)$ and the transport map $\pi=(\pi_{s \targ}) \in [0,1]^{(S \cup A_k) \times T}$:
\begin{equation}\label{eq:intro-reg-wasserstein}
\begin{aligned}
\RTW_\lambda(S\cup A_k)
&= \min_{\pi \ge 0,\, w}\quad 
    \underbrace{
        \sum_{s \in S\cup A_k}\sum_{\targ \in T}
        \pi_{s\targ}\,c(s,\targ)
    }_{{\color{biascolor}\text{bias}}}
    +
    \underbrace{
        \lambda \sum_{s \in S \cup A_k} w_s^2
    }_{{\color{regcolor}\text{regularization}}}
    \\[0em]
\text{s.t.}\quad &
    \forall t \in T, ~~ \sum_{s \in S\cup A_k}\pi_{st}
    = \frac{1}{|T|},
    \quad \text{and} \quad \forall s \in S\cup A_k,~
    \sum_{\targ \in T}\pi_{st}
    = w_s.
\end{aligned}
\end{equation}

\end{definition}
\end{tcolorbox}

Once we select the set $A_k$ to minimize the RTW objective, we then query the labels of $A_k$ and train on $S\cup A_k$ using the weights $w$ obtained from \eqref{eq:intro-reg-wasserstein}. 
The costs $c(s,t)$ measure some notions of dissimilarity, as discussed above. 
The objective function (\ref{eq:intro-reg-wasserstein}) captures two aspects of our problem: choosing the queries $A_k$, and also reweighting the augmented source points $S\cup A_k$ to be representative of $T$.  %

The two terms in the objective perform different roles. 
The first term in the objective is the optimal transport cost between the distribution $w$ supported on $S \cup A_k$ and the uniform distribution over the target dataset $T$. This captures the bias in using the distribution on $S \cup A_k$ given by $w$ to approximate the dataset $T$.  However, this could concentrate too much weight on only a few points, which decreases the effective sample size and causes large statistical error. 
The second term is a regularization term that prevents this large statistical error. 

Incorporating these two factors allows us to prove that a \emph{reweighted empirical risk minimizer} for the (reweighted) augmented training set $S \cup A_k, w$ will generalize to the test distribution. See Theorem \ref{thm:conditional-risk new} and Theorem~\ref{thm:transport-optimizer-recovery} in Section~\ref{sec:conditional-risk} for a formal statement that upper bounds the excess risk on the target distribution using an appropriate choice of the cost function $c$.

\paragraph{Algorithm with Provable Guarantees of Approximate Optimality.}  The generalization bound in Theorem~\ref{thm:conditional-risk new} with the RTW objective gives statistical guarantees for our approach. 
However, the computational challenge of minimizing the RTW problem \eqref{eq:intro-query-selection} efficiently remains.  For a fixed set \(A_k\) of queries, computing the RTW objective is a convex optimization problem.  The challenge is therefore to choose the query set \(A_k\).
Even in the case where $\lambda=0$, this corresponds to the classic $k$-median problem, which is NP-hard in the worst-case~\cite{ShmoysWilliamson}.  In general, when \(\lambda \neq 0\), the regularization term enforces some load balancing on the source weights.  This is similar to the capacitated \(k\)-medians problem, which is algorithmically even more challenging than the standard uncapacitated version~\citep{trinh2016survey}. %
On the other hand, as \(\lambda \to \infty\), this corresponds to assigning uniform weights across \(S \cup A_k\).  This is the setting considered by \cite{mahmood2022lowbudget}, who design heuristics to minimize this objective.  

Our main technical contribution is an efficient {\em polynomial time algorithm} that gives {\em provable guarantees for approximate optimality} for the RTW objective.
Our key insight is to prove that the function \eqref{eq:intro-reg-wasserstein} is \emph{supermodular} for every choice of \(\lambda \ge 0\). The unregularized objective when $\lambda=0$ is supermodular since the transport map $\pi$ just corresponds to the nearest neighbor assignment. When $\lambda>0$ the transport map is no longer the nearest neighbor assignment, yet surprisingly the objective remains supermodular. To the best of our knowledge, this was not known for the regularized version of the problem. %
This supermodularity property allows us to prove that a simple greedy algorithm that adds to the query set $A_k$ the point $\argmin_{\targ \in T} \RTW_\lambda(S \cup A_k \cup \{\targ\})$ achieves an approximate optimality guarantee. Hence the RTW problem \eqref{eq:intro-query-selection} provides a natural objective that achieves two important goals simultaneously: it allows us to obtain statistical guarantees for active domain adaptation (through regularization), while also ensuring computational tractability (through supermodularity).      

\paragraph{Validation.}  We implement our algorithm and validate it on standard domain adaptation benchmarks.  On benchmarks, we see that our method is competitive with state-of-the-art methods (including those that use information about the labels to choose their queries), and even outperforms them on some datasets. On the three tasks in Figure~\ref{fig:minidomainnet-three-to-real}, RTW outperforms all baselines at $k=50$ queries (out of a pool of 52,252 target samples), improving accuracy by 14--20 percentage points over training on the source alone. %

\section{Related Work}\label{sec:related}

We describe some of the other related work in this section. Please see Appendix \ref{sec:appendix-related-work} for a more detailed overview of related work.

\paragraph{Discrepancy and domain adaptation.}
In \emph{domain adaptation}, the learner is given access to labeled data from a source distribution \(\calP\) and unlabeled data from a target distribution \(\calQ\), with the goal of learning a model for \(\calQ\).   This is similar to our problem when the number of label queries \(k = 0\).
\citet{benDavid2010domains} give a generalization bound for a classifier learned on \(\calP\) when applied to \(\calQ\).  They bound this via a notion of \emph{discrepancy}, the \(\calH \Delta \calH\) measure, which captures how different \(\calP\) is from \(\calQ\), with respect to the hypothesis class \(\calH\).  Subsequent work defines a variety of other discrepancy measures that are tighter in certain regimes of the problem.  These discrepancy measures also capture lower bounds for domain adaptation-- when \(\calQ\) looks very different from \(\calP\) with respect to \(\calH\). %
The primary difference is that unlike standard domain adaptation, in active domain adaptation we are able to leverage additional label queries from $Q$, and many unlabeled samples from $Q$. These allow us to circumvent  some of the fundamental bottlenecks in domain adaptation.

\paragraph{Active learning.}  In \emph{active learning}, the goal of the learner is to learn a good hypothesis from a class \(\calH\) for a distribution \(\calQ\).  The learner is given \emph{unlabeled} samples from \(\calQ\), and can query the labels of a few points.  This is similar to our problem when the learner is given an empty training set \(S\).  A line of work beginning with \citet{dasgupta05} studies this problem in many settings, and characterizes the learnability of different hypothesis classes in a distribution-free setting~\citep{hanneke2014}.  Our work differs from the standard viewpoint of active learning in a few ways.  First, we have access to an additional training set \(S\), that may not be representative of \(\calQ\).  Thus our methods must adapt to and take advantage of this extra information.  %
Second, we design \emph{task agnostic} methods, that do not depend on the realized labels of the training set \(S\) or the queried points from \(T\).  This differs from traditional active learning methods, which usually depend on the labels.  

\paragraph{Task-agnostic active domain adaptation methods.}  
There are a few works that consider the problem of \emph{active domain adaptation}.  That is, they are given labeled samples from \(\calP\) and unlabeled samples from \(\calQ\).  Their goal is to query the labels of a few representative samples from \(\calQ\) and use this to learn a good classifier for \(\calQ\) with respect to the hypothesis class \(\calH\).  Our work focuses on designing \emph{task agnostic} methods.  That is, our methods choose which target points to query and find a reweighting of the augmented dataset, independently of the realized labels.  This gives us a general way to augment the training set, which can be used to train a downstream model for any number of tasks.  Prior work by \cite{chattopadhyay2013joint} designs a method to choose the queries and weights to minimize the \emph{maximum mean discrepancy} to \(\calQ\).  \citet{brunsSmithFeller2022balancing} choose weights that balance worst-case bias against weight dispersion.
\citet{deMathelin2022discrepancy} give a task-agnostic method based on \(k\)-medoids.  Their objective is essentially the \(\lambda = 0\) setting of our objective function.  They give an algorithm to approximately minimize this objective via submodular maximization.  Our method is similar, though establishing submodularity seems significantly less straightforward in the case where \(\lambda \neq 0\).    We provide more comparison to this method in the theory and experimental sections. The work of \citet{mahmood2022lowbudget} also provides a task-agnostic method that picks the next query based on the optimal transport cost; but they do not reweigh the distribution.  Our contribution is to frame the RTW objective, which provably ensures generalization, while also being computationally tractable.

\section{Regularized Reweighting Objective and Generalization Guarantees}
\label{sec:conditional-risk}
In this section, we describe a general, discrepancy-based objective function for the selection of points that depends on the hypothesis class and loss function. 
The objective is motivated by a novel generalization guarantee. 
Recall that the end goal for selecting the query points and performing reweighting is to learn a good hypothesis function using the augmented dataset. 

We begin with some new definitions and notation. 
From now on, we use index $i$ for features and labels from the augmented dataset $S\cup A_k$ and $j$ for the target dataset $T$. 
Let, for $h\in\mathcal{H}$, $g_h$ be the function defined as $g_h(x)=\Exp_{Y\sim Y|x}[\ell(h(x),Y)]$. 
For a hypothesis class $\mathcal{H}$ and loss function $\ell$, we define the following discrepancy measure  between two distributions: 
\begin{align}
\label{eq: def general discrepancy}
\operatorname{disc}_{\mathcal H,\ell}(\mathcal P_x ,\mathcal Q_x )
:= \sup_{h, h' \in\mathcal H}\Big| \mathbb E_{\mathcal P_x} [g_h(X) - g_{h'}(X)] - \mathbb E_{\mathcal Q_x} [g_h(X) - g_{h'}(X)] \Big|.
\end{align}
This measures the maximum error when estimating the difference in expected risks over $\mathcal Q_x$ of two hypotheses $h,h'$, if we instead estimate it on a different covariate distribution $\mathcal P_x$. 
This is a fine-grained variant of the $\mathcal Y$-discrepancy in~\citet{mohri2012drifting}, replacing a single $g_h$ by their pairwise differences, analogous to the classical notion of discrepancy~\citep{mansour2009domain}. 
It also equals the sum of the two directional transfer measures in~\citet{zhang2021transferability}.

Let $\nu_w$ be a distribution over the augmented source set $S_k = S\cup A_k$, where each $s_i\in S_k$ has weights $\nu_w(s_i) = w_i$ and label drawn from the labeling conditional distribution $Y_i\sim {Y}|s_i$. 
Let $\mu$ be the uniform empirical distribution over the target features $T$, and write $N=|T|$.
We distinguish the realized weighted training risk from the target population risk:
\[\widehat L_{\nu_w}(h) = \sum_{s_i\in S_k} w_i\ell(h(s_i),Y_i), \qquad L_Q(h)=\mathbb E_{(X,Y)\sim\mathcal Q}\ell(h(X),Y)=\mathbb E_{\mathcal Q_x}g_h(X).\]
The pool conditional risk $L_\mu(h)=N^{-1}\sum_{t\in T}g_h(t)$ will serve as an intermediate quantity in the analysis.
Let $\mathcal F_\ell=\{(x,y)\mapsto\ell(h(x),y):h\in\mathcal H\}$ denote the loss-composed hypothesis class, and let $\operatorname{Pdim}(\mathcal F_\ell)$ be its pseudo-dimension.
Our first result is a generalization bound for ERM on the realized weighted source distribution:
\begin{theorem}
\label{thm:conditional-risk new}

Suppose $A_k$ and $w$ are chosen using features only.
Conditional on all feature locations, assume the labels are independent under the shared conditional distribution $Y\mid x$.
Let $\hat h\in\arg\min_{h\in\mathcal H}\widehat L_{\nu_w}(h)$ be a minimizer of the weighted empirical risk on augmented source $S_k$.
If $0\leq\ell\leq1$ and $\operatorname{Pdim}(\mathcal F_\ell)<\infty$, then  there is a universal constant $C$ such that, for every $\delta\in(0,1)$, with conditional probability at least $1-\delta$ over the labels,
\begin{align}
\label{eq: generalization guarantee}
 L_Q(\hat h)-\inf_{h\in\mathcal H} L_Q(h)\leq
\operatorname{disc}_{\mathcal H,\ell}(\nu_w, \mathcal Q_x)+
C\|w\|_2\left(\sqrt{\operatorname{Pdim}(\mathcal F_\ell)}+\sqrt{\log\frac2\delta}\right).
\end{align}
\end{theorem}

The upper bound in equation
(\ref{eq: generalization guarantee}) admits an intuitive decomposition: the first discrepancy term is a bias term, measuring the distortion in target population risk differences when using the weighted training features instead.
The second term is a variance term, coming from the realistic modeling assumption that the label is not a deterministic function of the feature.

\subsection{Regularized Transport Weighting}
The discrepancy term can be upper bounded by a transportation problem.
For $s,t\in \mathcal{X}$, define cost $c$ to satisfy:
\begin{align}
    \label{eq: def of general dissimilarity}
    c(s,t) \geq \sup_{h,h'\in \mathcal{H}}\left|g_h(s)-g_{h'}(s)-g_h(t)+g_{h'}(t)\right|.
\end{align}
Intuitively, to compare the expected losses of hypotheses $h$ and $h'$ on point $t\in \calX$, then $c(s, t)$ is the maximum amount of error we get if we compare their expected losses on $s$ instead. 
Note here that if all functions $g_h$ are $L$-Lipschitz and $\mathcal{X}$ is a metric space, then this cost can be taken to be $c(s,t) = 2L\cdot d(s,t)$.

The triangle inequality gives $\operatorname{disc}_{\mathcal H,\ell}(\nu_w,\mathcal Q_x)\leq\operatorname{disc}_{\mathcal H,\ell}(\nu_w,\mu)+\operatorname{disc}_{\mathcal H,\ell}(\mu,\mathcal Q_x)$.
The second term is independent of the queries and weights. Bounding the first by transport cost and incorporating the variance term in Equation~\ref{eq: generalization guarantee} gives the following optimization problem:
\begin{equation}
\label{eq:unsquared-objective}
\min_{\pi,w\geq 0} \sum_{s \in S_k} \sum_{t \in T} \pi_{st} c(s, t) + \lambda_0\|w\|_2,
\end{equation} 
where $\sum_{s\in S_k}\pi_{st} = \frac{1}{|T|}\text{ for all }t\in T$, and $w_s = \sum_{t\in T} \pi_{st}$ for all $s$. 

By replacing the $\ell_2$ regularization $\lambda_0 \|w\|_2$ with the \textit{squared} $\ell_2$ regularization term $\lambda \|w\|_2^2$, we obtain our $\RTW_\lambda$ objective presented in Equation \ref{eq:intro-reg-wasserstein}.
The squared $\ell_2$ regularization has better properties (see Section~\ref{sec:algorithm}), and an appropriate choice of $\lambda$ also controls the excess population risk, up to a target-pool sampling term.
This is formalized by the following theorem. 
\begin{theorem}
\label{thm:transport-optimizer-recovery}
Under the assumptions of Theorem~\ref{thm:conditional-risk new}, suppose $T$ consists of $N\geq1$ i.i.d. samples from $\mathcal Q_x$.
Let $S\cup A_k$ be nonempty and fix a cost $c$ satisfying~\eqref{eq: def of general dissimilarity}.
For $\delta\in(0,1)$, take $\lambda_0=C\left(\sqrt{\operatorname{Pdim}(\mathcal F_\ell)}+\sqrt{\log(4/\delta)}\right)$, with $C$ as in Theorem~\ref{thm:conditional-risk new}.
Let $B$ be the minimum of~\eqref{eq:unsquared-objective}, attained at $(\pi_0^*,w_0^*)$ chosen using features only, and set $\lambda=\lambda_0/(2\|w_0^*\|_2)$.
Let $(\pi^*,w^*)$ be a minimizer of $\RTW_\lambda(S\cup A_k)$ chosen using features only, and let $\hat h\in\arg\min_{h\in\mathcal H}\widehat L_{\nu_{w^*}}(h)$.
There is a universal constant $C'$ such that, with probability at least $1-\delta$ over the sampled features and labels,
\[
L_Q(\hat h)-\inf_{h\in\mathcal H}L_Q(h)
\leq 2\RTW_\lambda(S\cup A_k)
+\frac{C'}{\sqrt N}\left(\sqrt{\operatorname{Pdim}(\mathcal F_\ell)}+\sqrt{\log\frac4\delta}\right).
\]
\end{theorem}
The proof is in Appendix~\ref{app:squared-regularization}.

For a specified $L$, the Wasserstein term uses only the feature locations and weights, without requiring the conditional risks themselves.
The proofs of Theorems~\ref{thm:conditional-risk new} and~\ref{thm:transport-optimizer-recovery} appear in Appendix~\ref{app:conditional-risk-proofs}.

\section{Our Algorithm and Approximate Optimality Guarantees}\label{sec:algorithm}

In this section, we fix $\lambda \geq 0$ and study the computational aspects of the Regularized Transport Weighting problem in \eqref{eq:intro-query-selection}: given $S,T,k$, the costs $c$ and $\lambda\geq0$, find $A_k \subset T$ with $|A_k|=k$ that minimizes $\RTW_\lambda(S \cup A_k)$.

Recall that for a fixed set of query points $A_k \subset T$, the quantity $\RTW_\lambda(S \cup A_k)$ is the value of a convex program given in \eqref{eq:intro-reg-wasserstein}. The convex program is a regularized transport problem where the distribution supported on $S \cup A_k$ can be reweighted.  This is related to the Wasserstein objective in \citep{mahmood2022lowbudget}; crucially however, we allow the distribution on the source to be reweighted (as opposed to uniform weights). While the {\em bias } term in our objective \eqref{eq:intro-reg-wasserstein} captures the transport cost (or Wasserstein distance) between the target and the reweighted distribution $w$ on $S \cup A_k$, the regularization term plays an important role in obtaining generalization in domain adaptation (e.g., with stochastic $Y|x$).  The convex program \eqref{eq:intro-reg-wasserstein} can be solved efficiently in polynomial time; moreover in Section~\ref{sec:experiments} we describe a dual-based algorithm that is more efficient in practice.

The primary computational challenge is in identifying a good choice of $A_k$ that minimizes objective \eqref{eq:intro-query-selection}. The problem is NP-hard even in the special case when $\lambda=0$. The regularized transport weighting objective interpolates between two familiar regimes.
When $\lambda=0$, it reduces to $k$-median with $S$ as pre-opened centers and $A_k$ as the $k$ additional centers, recovering $k$-medoids when $S=\varnothing$. For $\lambda \in (0, \infty)$, it is analogous to the more challenging capacitated $k$-median~\citep{trinh2016survey}, except it allows fractional assignments and incentivizes a spread load instead of constraining the max load. As $\lambda\to\infty$, the weights become uniform, and the objective converges to the uniformly weighted Wasserstein selection criterion of \citet{mahmood2022lowbudget}. 

\subsection{Greedy querying algorithm}

\begin{figure}[t]
    \centering
    \includegraphics[width=0.85\linewidth]{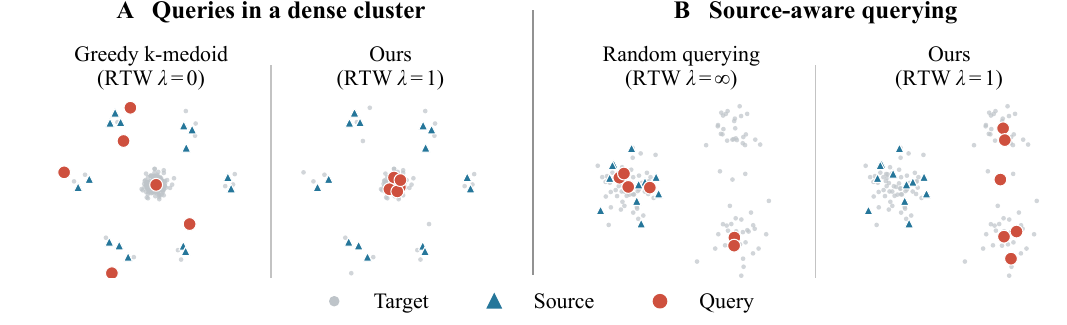}
    \caption{Illustration of our query selection method (Section \ref{sec:algorithm}) on two synthetic instances. When $\lambda=0$ our algorithm reduces to the (greedy) $k$-medoids algorithm~\citep{deMathelin2022discrepancy}. For intermediate values of $\lambda$, our algorithm greedily optimizes $\RTW_\lambda$, overcoming the failures of both greedy $k$-medoids and random sampling. 
    }
    \label{fig:query-method-comparison}
    \vspace{-10pt}
\end{figure}
We propose a greedy algorithm (Algorithm (\ref{alg:rtw_algorithm})) for solving problem~\eqref{eq:intro-query-selection}. We run for $k$ iterations: at each iteration we select the point that brings down the objective $\RTW_\lambda$ the most, and add it to our set $A_k$. See Figure~(\ref{fig:query-method-comparison}) to compare how our greedy RTW algorithm differs from the (greedy) $k$-medoids algorithm and random sampling in the way it picks the query points.

\begin{algorithm}[t]
    \caption{Greedy Regularized Transport Weights}\label{alg:rtw_algorithm}
    \begin{algorithmic}
        \STATE \textbf{Inputs:} source points $S \subseteq \mathcal{X}$, target points $T \subseteq \calX$ and query budget $k$. 
        \STATE \textbf{Hyperparameters: } $\lambda \in \R_{+}:$ regularization strength
        \STATE \textbf{Outputs: } $A_k \subseteq T$: the set of points to query for labels,
        $w^* \in \Delta(S \cup A_k)$: weights of the augmented training set.
    \end{algorithmic}
        
    \begin{algorithmic}[1]
        \STATE Initialize $A_0 = \emptyset$, $i=0$.
        \WHILE{$i < k$}
        \STATE Compute $x_i^* = \argmin_{x \in T \setminus A_i} \ReWD_\lambda(S \cup A_i \cup \{x\})$
        \STATE $A_{i+1} \gets A_i \cup \{x_i^*\}$
        \STATE $i \gets i+1$
        \ENDWHILE
        \STATE Set $w^*$ to be the optimum weights $w$ for $\RTW_\lambda(S \cup A_k)$ defined in \eqref{eq:intro-reg-wasserstein}.  %
        \STATE Output $A_k, w^*$.
    \end{algorithmic}
     
\end{algorithm}

\subsection{Algorithmic Guarantees}

Algorithm (\ref{alg:rtw_algorithm}) outputs a set $A_k$ in $k$ iterations in polynomial time. Our main theoretical contribution is to show that Algorithm~(\ref{alg:rtw_algorithm}) outputs a set that is approximately optimal--- it brings down the cost function by at least a constant times the best possible subset. 
The proof is given in Appendix~\ref{app:greedy-guarantee}.

\begin{theorem}\label{thm:algorithmic_guarantee}
Suppose $\lambda\ge 0$, $S\ne\varnothing$, and $1\leq k\leq |T|$ is an integer. Let $A^*_{k} = \argmin_{A \subseteq T, |A| = k} \RTW_\lambda(S \cup A)$. Then, Algorithm (\ref{alg:rtw_algorithm}) given query budget $k$ outputs a set $\hat A_{k}$ in polynomial time satisfying
    \begin{equation*}
        \RTW_\lambda(S) - \RTW_\lambda(S \cup \hat A_k) \ge \inparen{1 - \frac{1}{e}}\inparen{\RTW_\lambda(S) - \RTW_\lambda(S \cup A^*_{k})}
    \end{equation*}

Moreover, for $\epsilon,\eta\in(0,1)$, suppose $\RTW_\lambda(S \cup A^*_k) \le \eta \cdot \RTW_\lambda(S)$. Then our algorithm with query budget $k'=\min\{|T|,\lceil k\log(1/(\epsilon\eta))\rceil\}$ outputs $\hat{A}_{k'}$ such that
\[\RTW_\lambda(S \cup \hat{A}_{k'}) \le (1+\epsilon)\cdot  \eta \RTW_\lambda(S).\]

\end{theorem}

The crucial insight behind our analysis is the surprising {\em supermodularity} property of the function $\RTW_\lambda(S)$. For fixed $S$, the decrease $\RTW_\lambda(S) - \RTW_\lambda(S \cup A)$ is a monotone submodular function of $A$. The corresponding diminishing-returns inequality is: for every $S \subseteq S', x \not\in S'$,
 \begin{equation*}
         \RTW_\lambda(S) - \RTW_\lambda(S \cup \{x\}) \ge \RTW_\lambda(S') - \RTW_\lambda(S' \cup \{x\}).
 \end{equation*}

See Theorem~\ref{thm:supermodularity_rtw} for the formal claim in Appendix~\ref{sec:appendix-algorithm}. While the supermodularity for $\lambda=0$ is easy to verify (the transport map is just the nearest neighbor assignment), it becomes more challenging for $\lambda>0$. The proof of the submodularity claim for the general regularized objective is new to the best of our knowledge.

The proof of submodularity involves a flow-based proof. A key change in perspective involves the function $f_S(q)$ for any non-negative vector of demands $q \in \mathbb{R}^T_{\ge0}$, which corresponds to the optimal regularized transport cost of matching the ``target vector'' $q$ and allows us to better control the change when a new query point is added. 
We remark that the same proof also works when the regularizer $\lambda \norm{w}_2^2$ is replaced by any separable sum $\lambda \sum_i \phi(w_i)$ where $\phi$ can be an arbitrary non-negative convex function.

\section{Experiments}
\label{sec:experiments}
To evaluate our algorithm, we conducted experiments on two well known image datasets used for domain adaptation, OfficeHome-RSUT (\citet{tan2020officehome_RSUT}) and mini-DomainNet (\citet{zhou2021dael}). We also conducted some experiments on synthetic datasets, designed to illustrate situations where the baselines can fail, while our method still succeeds. 

\paragraph{RTW Implementation.} Algorithm (\ref{alg:rtw_algorithm}) involves solving multiple quadratic optimization problems subject to linear constraints. Our implementation solves these using gradient ascent on the unconstrained dual, which we found to be the most stable approach in our experiments. To further improve computational efficiency, we solve the optimization problems only for a randomly sampled subset of the candidate points. Further implementation details can be found in the appendix.

\paragraph{Baselines.} We compare the performance of our proposed algorithm with that of other established techniques to mitigate the distribution shift between the source and the target.  Broadly,  we consider two types of baselines. Importance-weighting methods use labeled source samples and unlabeled target samples to estimate importance weights for the source data. A classifier is then trained on the labeled source samples using these weights. Active domain adaptation methods may additionally query the labels for a few selected target points, and use them during classifier training. 

We consider three importance weighting baselines - \textbf{Kernel Mean Matching (KMM)} (\citet{huang2007selectionBias}), \textbf{ KLIEP} (\citet{sugiyama2007kliep}) and \textbf{ Importance Weighting Network (IWN)} (\citet{demathelin2022iwn}). We consider five active domain adaptation methods - \textbf{K-medoids, K-medoids+W} (\citet{deMathelin2022discrepancy}), \textbf{MaxHerd} (\citet{bae2024maxherd}), \textbf{BADGE} (\citet{ash2020badge}) and \textbf{CLUE} (\citet{prabhu2021clue}). \cite{deMathelin2022discrepancy} consider two weighting schemes - uniform weights which assign equal weight to all training samples, and balanced weights, which reweigh the points so that the original source and queried target samples contribute equally. In our experiments, we found that balanced reweighting performs best for all baselines, and so all reported results use balanced reweighting. For detailed descriptions of the baselines, see Appendix \ref{sec:appendix-experiments}.

\subsection{Experimental Setup and Results}

\paragraph{OfficeHome-RSUT.} OfficeHome-RSUT, introduced by \citet{tan2020officehome_RSUT}, is a class-imbalanced variant of the original OfficeHome dataset \citep{venkateswara2017officehome}, an image dataset for domain adaptation. It consists of images from 3 domains - Real World, Clipart and Product. Each domain consists of images belonging to one of 65 object categories. We consider six tasks, corresponding to all possible source and target domains. 

We use an ImageNet pre-trained ResNet-50 network \citep{he2016resnet} to extract 2,048-dimensional, $\ell_2$-normalized embeddings of all the images in the dataset. The embeddings remain frozen throughout our experiments and we train only a linear classification head. For each method, we train a logistic-regression classifier on the labeled source samples together with any queried target samples, with the weighting prescribed by the method.

For each  task, we generate five random partitions of the target dataset, using 80\% of the samples as the acquisition pool and keeping 20\% of the samples as a held-out evaluation set. All methods use the same partition, and each method is only allowed access to the samples in the acquisition pool. Each method is evaluated by the performance on the held-out samples. We set the query budget $K = 100$, and evaluate accuracy on the held-out set after every five queries.

\begin{figure}[t]
    \centering
    \includegraphics[width=\linewidth]{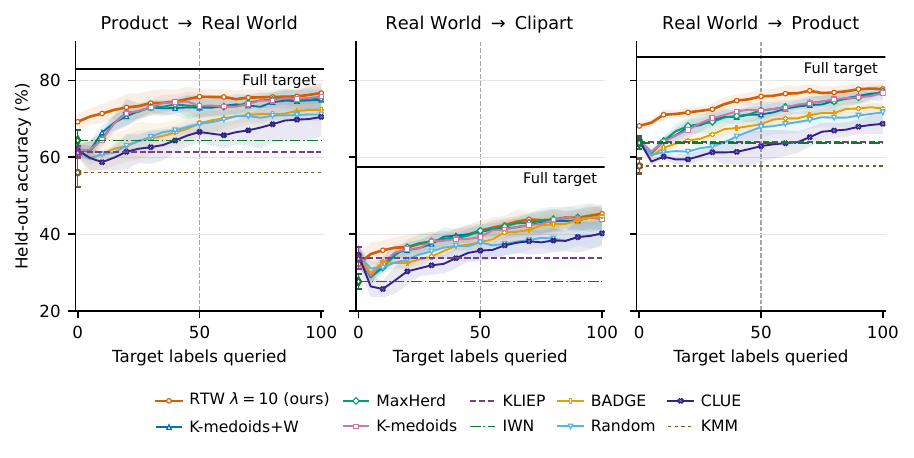}
\vspace{-25pt}    
    \caption{
    Results for three tasks on OfficeHome-RSUT. The full target line denotes the accuracy of the classifier trained on the target query pool. Dashed lines represent methods that do not query for additional labels.
    Our method outperforms existing methods, and with only a few additional queries we are able to improve the accuracy of the trained classifier significantly. 
    }\label{fig:officehome_rsut_main}
    \vspace{-10pt}
\end{figure}

\noindent{\textit{Results: }} Figure (\ref{fig:officehome_rsut_main}) shows the accuracies evaluated at different query budgets on three selected tasks. At $K = 100$ queries, RTW achieves an accuracy of $76.73 \pm 2.61\%$ on Product$\to$ Real World and $45.29 \pm 2.47 \%$ on Real World $\to$ Clipart, as compared to the best-performing baseline's accuracy of $75.86 \pm 3.45 \%$ and $44.61 \pm 2.19 \%$, achieved by K-medoids and BADGE respectively. In addition, RTW also outperforms other algorithms at lower query budgets. In particular, reweighting the source with our method outperforms existing importance weighting baselines even without additional queries. Additional results on all tasks can be found in Appendix \ref{sec:appendix-experiments}.

\paragraph{mini-DomainNet.}
mini-DomainNet is a subset of DomainNet containing 126 object categories across four domains introduced by~\citet{zhou2021dael}: Clipart, Painting, Sketch, and Real.
We evaluate all 12 directed adaptation tasks across these four domains. Figure~\ref{fig:minidomainnet-three-to-real} shows the three tasks with Real as the target domain, and figures for the remaining nine can be found in Figures~\ref{fig:minidomainnet-to-clipart}--\ref{fig:minidomainnet-to-sketch} in the appendix.
After removing duplicate images and samples with conflicting labels, we retain all source images. As in our OfficeHome-RSUT experiments, we construct five  partitions of the target domain into a query pool using 80\% of the images, and maintain a held-out evaluation set consisting of the remaining 20\% of the images.

As in OfficeHome-RSUT, we use frozen, 2,048-dimensional, $\ell_2$-normalized ResNet-50 ImageNet-V1 features, and train a weighted logistic regression on the labeled source samples augmented by the queried target samples.
We set a query budget of $k = 100$, and compute accuracy on the evaluation set every five queries.

\noindent{\textit{Results: }}
At $K=100$, RTW with $\lambda=10$ achieves $72.87 \pm 0.29\%$ on Painting$\to$Real and $66.19 \pm 0.28\%$ on Clipart$\to$Real, exceeding the strongest baseline, $K$-medoids+W ($68.83 \pm 0.43\%$ and $63.95 \pm 0.52\%$, respectively); its advantage is already present at $K=0$ through source reweighting.
On Sketch$\to$Real, RTW reaches $63.05 \pm 0.41\%$, below $K$-medoids+W ($64.62 \pm 0.48\%$), $K$-medoids ($64.45 \pm 0.62\%$), and MaxHerd ($64.38 \pm 0.63\%$).
On the remaining nine tasks, RTW outperforms all query baselines at both $K=50$ and $K=100$ except on Painting$\to$Sketch (Appendix~\ref{sec:appendix-experiments}). 

\paragraph{Synthetic Experiments.} We also conducted experiments on synthetic datasets, to illustrate simple settings where our algorithm has a natural advantage over many other popular methods in the query selection. These synthetic tasks were regression tasks. Intuitively, the dataset consists of a diffuse cluster with noiseless labels, along with some dense clusters with noisy labels. While $k$-medoids spends most of its queries covering the diffuse cluster, RTW allocates more queries to the noisy locations. The full details can be found in Section \ref{sec:synthetic_experiments}.

\section{Conclusion}
We provide a general and principled task-agnostic approach for active domain adaptation. Our notion of regularized transport weights (RTW) suggests new statistically tractable ways to leverage the geometry of the hypothesis class to compute weights between the training data and test dataset.  %
We provide efficient algorithms to approximately minimize this objective, with provable guarantees for approximate optimality via the toolkit of submodular function maximization.  This gives a simple and practical method that competes with the state of the art on a variety of applications.

\section*{AI Disclosure}

In this work, we used generative AI tools in generating synthetic datasets, providing proof ideas for mathematical claims, experimental methodology, and implementing methods. We have not used generative AI tools for 
developing theoretical models, formulating mathematical claims, proposing and refining hypotheses, or interpreting results,  
and translation and qualitative data analysis are not applicable to this work.  We have reviewed all AI-assisted work.  All proofs have been verified and written by the authors.  We take responsibility for the final content of this work, including text, claims or artifacts produced with the aid of generative AI.

\subsubsection*{Acknowledgments}
We gratefully acknowledge the support of the NSF-Simons AI-Institute for the Sky (SkAI) via grants NSF AST-2421845 and Simons Foundation MPS-AI-00010513. Guo, Sinha, Arvanitakis, Srinivas (while a student at Northwestern) and Vijayaraghavan were also partly funded by NSF ECCS 2216970. 
Ji acknowledges support from the Alfred P. Sloan Research Fellowship.

Fermilab is managed by FermiForward Discovery Group, LLC under Contract No. 89243024CSC000002 with the U.S. Department of Energy, Office of Science, Office of High Energy Physics. The United States Government retains and the publisher, by accepting the article for publication, acknowledges that the United States Government retains a non-exclusive, paid-up, irrevocable, world-wide license to publish or reproduce the published form of this manuscript, or allow others to do so, for United States Government purposes.

\bibliography{references}
\bibliographystyle{iclr2027_conference}

\clearpage
\appendix

\tableofcontents
\section{Extended Related Work}
\label{sec:appendix-related-work}

\paragraph{Discrepancy and domain adaptation.}
\citet{benDavid2010domains} use the $\mathcal H\Delta\mathcal H$ divergence to obtain a domain-adaptation generalization bound.
\citet{mansour2009domain} extend the analysis to general losses and motivate source reweighting.
Source-guided discrepancy fixes one hypothesis in the classical pairwise discrepancy to a source-risk minimizer, using source labels without requiring target labels~\citep{kuroki2019sourceguided}.
Localized discrepancy instead restricts attention to hypotheses with small source risk~\citep{zhang2020localized}.
Generalized discrepancy allows the source weights to depend on the hypothesis rather than fixing one weighting for the entire class~\citep{cortes2019generalized}.
For multiclass adaptation, \citet{zhang2019bridging} introduce margin disparity discrepancy and use it to derive an adversarial learning algorithm.
Connecting discrepancy to label acquisition, \citet{deMathelin2022discrepancy} use a label-consistent localized class to obtain single-shot active domain adaptation bounds and a geometric $K$-medoids query rule.

\citet{redko2017transport} obtain Wasserstein-based adaptation bounds, including a setting with both source and target labels.
The $\mathcal Y$-discrepancy of \citet{mohri2012drifting} measures the largest absolute risk difference over a hypothesis class between two joint input--label distributions, allowing stochastic labels and general bounded losses.
\citet{zhang2021transferability} instead compare excess risks; under covariate shift and a common hypothesis class, our conditional-risk discrepancy equals the sum of their two directional transfer measures.

\paragraph{Balancing weights.}
Kernel mean matching corrects sample-selection bias by matching weighted source and target feature means, without estimating either density~\citep{huang2007selectionBias}.
\citet{yu2012kmm} give high-probability bounds for this estimator under regularity conditions on the regression function and kernel.
\citet{brunsSmithFeller2022balancing} balance squared IPM bias against weight dispersion; for a uniform empirical source, the dispersion term is a squared $\ell_2$ penalty.
Their weights are placed on a fixed support, while our support $S\cup A_k$ changes with the selected target batch.

Weights can also be learned together with a predictor: WANN uses adversarial training with a $\mathcal Y$-discrepancy objective to learn a regressor and weights over labeled source and target samples~\citep{deMathelin2021wann}.
Best-Effort Adaptation provides discrepancy-based bounds that hold uniformly over the weights used to combine labeled source and target data~\citep{awasthi2024besteffort}.
Both take the labeled target sample as given; we also choose which target labels to acquire.
Weights play a different role in importance-weighted active learning (IWAL): inverse query-probability weights correct the bias introduced by sequential active sampling~\citep{beygelzimer2009iwal}, whereas our weights are chosen by distribution matching for a single-shot batch.

\paragraph{Active domain adaptation.}
JO-TAL jointly reweights source samples and selects a target batch through an MMD-based convex relaxation~\citep{chattopadhyay2013joint}.
We additionally optimize the queried samples' training weights and penalize weight concentration.

AADA combines adversarial adaptation with acquisition scores based on predictive uncertainty and domain-discriminator outputs~\citep{su2020aada}.
TQS uses committee disagreement, uncertainty, and domainness, with randomized selection for diversity~\citep{fu2021tqs}.
CLUE incorporates diversity through uncertainty-weighted clustering of target embeddings~\citep{prabhu2021clue}.
EADA uses energy-based scores to account for both domain shift and instance uncertainty~\citep{xie2022eada}.
To address uncertainty miscalibration under shift, DUC uses Dirichlet predictions and a two-round selection rule~\citep{xie2023duc}.
Our acquisition rule instead uses only feature geometry once the representation is fixed.

\paragraph{Batch active learning.}
\citet{chattopadhyay2012marginal} select batches by matching the marginal distributions of labeled and unlabeled data.
\citet{viering2019nuclear} analyze single-shot selection through MMD and discrepancy bounds and introduce nuclear discrepancy under an optimistic probabilistic analysis.
\citet{shui2020deep} use a Wasserstein-based objective while alternating neural-network training and batch querying.

The core-set approach of \citet{sener2018active} selects batches by geometric coverage in a learned representation.
BADGE combines uncertainty and diversity by selecting diverse gradient embeddings~\citep{ash2020badge}.
iRDM selects regression samples without labels by balancing representativeness and diversity~\citep{liu2020irdm}.
DUAL learns an encoder--decoder representation and selects samples that preserve input patterns and cluster structure without supervision~\citep{li2020dual}.
We combine feature-only batch selection with adaptation from a source sample and regularized training weights.

\paragraph{Coresets and distribution matching.}
For logistic regression, \citet{munteanu2018logistic} construct sensitivity-sampling coresets under a data-complexity condition, approximating an already labeled training objective.
\citet{borsos2024bilevel} formulate weighted coreset construction as cardinality-constrained bilevel optimization of a model-training objective.
\citet{axiotis2024sensitivity} combine clustering and sensitivity sampling to approximate average loss under H\"older continuity in an embedding.
Our acquisition objective uses feature geometry rather than observed training losses, since queries must be selected before their labels are available.

Distributional coresets approximate a measure rather than a particular training objective.
Kernel thinning compresses a sample while controlling worst-case integration error over an RKHS~\citep{dwivedi2021thinning}.
Wasserstein measure coresets use transport distance to control families of downstream objectives~\citep{claici2018wasserstein}.
\citet{yin2025sinkhorn} optimize support locations with a Sinkhorn loss; their entropic transport regularization differs from our penalty on training weights.
Closer to our query-selection problem, \citet{mahmood2022lowbudget} select uniformly weighted batches by discrete Wasserstein minimization via integer programming.
We instead learn regularized weights over both source and queried target points.

\section{Proofs of the Risk Bounds}
\label{app:conditional-risk-proofs}

\subsection{Proof of Theorem~\ref{thm:conditional-risk new}}
\begin{proof}
Condition on the feature locations in $S$ and $T$, and on the feature-only choice of $A_k$ and $w$.
All probabilities and expectations over training labels below are conditional on these features and choices.
Write $S_k=S\cup A_k$ and $w_i=\nu_w(s_i)$ for $s_i\in S_k$, so $\sum_i w_i=1$.
The training loss is $\widehat L_{\nu_w}(h)=\sum_i w_i\ell(h(s_i),Y_i)$, while the population risk is $L_Q(h)=\mathbb E_{\mathcal Q_x}g_h(X)$.
Conditional on the features, the labels $Y_i$ are independent with the shared conditional law $Y\mid s_i$.
Write $Z_h=\sum_i w_i\{\ell(h(s_i),Y_i)-g_h(s_i)\}$.
For any $h,h'\in\mathcal H$, the definition of discrepancy and the triangle inequality give
$$\left|[L_Q(h)-L_Q(h')]-[\widehat L_{\nu_w}(h)-\widehat L_{\nu_w}(h')]\right|
\leq\operatorname{disc}_{\mathcal H,\ell}(\nu_w,\mathcal Q_x)+2\sup_{h\in\mathcal H}|Z_h|.$$
We first bound the noise term uniformly over $\mathcal H$.

\paragraph{Symmetrization.}
Let $Y_i'$ be independent copies of $Y_i$ from the same conditional distributions, independent across $i$, and let $\sigma_i$ be independent Rademacher random variables.
Since $g_h(s_i)=\mathbb E_{Y_i'}\ell(h(s_i),Y_i')$, using Jensen's inequality and triangle inequality, we have
$$\begin{aligned}
\mathbb E_Y\sup_h|Z_h|
&\leq\mathbb E_{Y,Y'}\sup_h\left|\sum_i w_i\{\ell(h(s_i),Y_i)-\ell(h(s_i),Y_i')\}\right|\\
&=\mathbb E_{Y,Y',\sigma}\sup_h\left|\sum_i w_i\sigma_i\{\ell(h(s_i),Y_i)-\ell(h(s_i),Y_i')\}\right|\\
&\leq2\mathbb E_{Y,\sigma}\sup_h\left|\sum_i w_i\sigma_i\ell(h(s_i),Y_i)\right|.
\end{aligned}$$

\paragraph{Weighted complexity bound.}
Conditional on the labels, consider the stochastic process $R_f=\sum_i w_i\sigma_i f(s_i,Y_i)$ indexed by functions $f\in\mathcal F_\ell$, and define the probability measure
$$Q_w=\frac1{\|w\|_2^2}\sum_i w_i^2\delta_{(s_i,Y_i)}.$$
Define $\|f-f'\|_{L_2(Q_w)}=\big(\mathbb E_{(x,y)\sim Q_w}[(f(x,y)-f'(x,y))^2]\big)^{1/2}$.
For every $s\in\mathbb R$,
$$\begin{aligned}
\mathbb E_\sigma\exp\{s(R_f-R_{f'})\}
&=\prod_i\mathbb E_{\sigma_i}\exp\{sw_i\sigma_i[f(s_i,Y_i)-f'(s_i,Y_i)]\}\\
&=\prod_i\cosh\!\left(sw_i[f(s_i,Y_i)-f'(s_i,Y_i)]\right)\\
&\leq\exp\left\{\frac{s^2}{2}\sum_i w_i^2[f(s_i,Y_i)-f'(s_i,Y_i)]^2\right\}\\
&=\exp\left\{\frac{s^2\|w\|_2^2}{2}\|f-f'\|_{L_2(Q_w)}^2\right\}.
\end{aligned}$$
The first equality uses independence; the second uses the equal probabilities of the two signs and $\cosh v=(e^v+e^{-v})/2$.
The last equality follows from the definition of $Q_w$, giving the sub-Gaussian increment bound.
The inequality $\cosh v\leq e^{v^2/2}$ follows by comparing power series: since $(2k)!\geq2^k k!$,
$$\cosh v=\sum_{k=0}^\infty\frac{v^{2k}}{(2k)!}
\leq\sum_{k=0}^\infty\frac{v^{2k}}{2^k k!}=e^{v^2/2}.$$

Let $\mathcal N(\epsilon,\mathcal F_\ell,L_2(Q))$ be the covering number at radius $\epsilon$ under $L_2(Q)$.
The pseudo-dimension covering bound for a class in $[0,1]$ gives, for every probability measure $Q$,
$$\log\mathcal N(\epsilon,\mathcal F_\ell,L_2(Q))\leq C_0\operatorname{Pdim}(\mathcal F_\ell)\log\frac{C_0}{\epsilon},\qquad 0<\epsilon\leq1,$$
where $C_0$ is universal \citep{anthony1999neural}.
A cover of $\mathcal F_\ell$ and its negation covers $\mathcal F_\ell\cup(-\mathcal F_\ell)$ with at most twice as many balls.
Applying the Dudley entropy integral to this signed class therefore gives
$$\begin{aligned}
\mathbb E_\sigma\sup_{f\in\mathcal F_\ell}|R_f|
&\leq C_1\|w\|_2\left[1+\int_0^1\sqrt{\log\left(2\mathcal N(\epsilon,\mathcal F_\ell,L_2(Q_w))\right)}\,d\epsilon\right]\\
&\leq C_2\|w\|_2\left(\sqrt{\operatorname{Pdim}(\mathcal F_\ell)}+1\right).
\end{aligned}$$
The constant term accounts for the coarse scale of the signed class, whose $L_2(Q_w)$-diameter is at most two.
The integral is finite because $\int_0^1\sqrt{\log(C_0/\epsilon)}\,d\epsilon<\infty$.
This bound holds for every realization of the labels, so symmetrization implies
$$\mathbb E_Y\sup_h|Z_h|\leq2C_2\|w\|_2\left(\sqrt{\operatorname{Pdim}(\mathcal F_\ell)}+1\right).$$

\paragraph{Concentration over labels.}
Changing one label $Y_i$ changes $\sup_h|Z_h|$ by at most $w_i$, since the loss lies in $[0,1]$.
McDiarmid's inequality consequently gives
$$\Pr\left\{\sup_h|Z_h|-\mathbb E_Y\sup_h|Z_h|\geq s\right\}\leq\exp\left(-\frac{2s^2}{\|w\|_2^2}\right).$$
Taking $s=\|w\|_2\sqrt{\tfrac12\log(1/\delta)}$ bounds the centered deviation with probability at least $1-\delta$.
Since $\sqrt{\log(2/\delta)}$ absorbs both the constant term and $\sqrt{\log(1/\delta)}$, we obtain, for a universal $C$,
$$\sup_h|Z_h|\leq C\|w\|_2\left(\sqrt{\operatorname{Pdim}(\mathcal F_\ell)}+\sqrt{\log\frac2\delta}\right).$$
For every $h\in\mathcal H$, the initial pairwise bound and the optimality of $\hat h$ give
$$\begin{aligned}
L_Q(\hat h)-L_Q(h)
&\leq\widehat L_{\nu_w}(\hat h)-\widehat L_{\nu_w}(h)+\operatorname{disc}_{\mathcal H,\ell}(\nu_w,\mathcal Q_x)+2\sup_{h'\in\mathcal H}|Z_{h'}|\\
&\leq\operatorname{disc}_{\mathcal H,\ell}(\nu_w,\mathcal Q_x)+2\sup_{h'\in\mathcal H}|Z_{h'}|.
\end{aligned}$$
Taking the supremum over $h$ on the left and absorbing the factor of two into the universal constant proves the theorem; no target-risk minimizer need exist.
\end{proof}

\subsection{Risk Bound via Squared Regularization}
\label{app:squared-regularization}

\begin{proof}[Proof of Theorem~\ref{thm:transport-optimizer-recovery}]
Use $B$, $(\pi_0^*,w_0^*)$, and $\lambda=\lambda_0/(2\|w_0^*\|_2)$ as in the statement.
For every feasible $(\pi,w)$, subtracting the unsquared objective from the squared objective and substituting this choice of $\lambda$ gives
\[
\lambda\|w\|_2^2-\lambda_0\|w\|_2
=\lambda\big(\|w\|_2^2-2\|w_0^*\|_2\|w\|_2\big).
\]
Adding the constant $\lambda\|w_0^*\|_2^2$ completes the square:
\[
\lambda\big(\|w\|_2^2+\|w_0^*\|_2^2-2\|w_0^*\|_2\|w\|_2\big)
=\lambda\big(\|w\|_2-\|w_0^*\|_2\big)^2\geq0.
\]
The two objectives therefore satisfy
\[
\begin{aligned}
&\underbracket{\Big(\sum_{s,t}\pi_{st}c(s,t)+\lambda\|w\|_2^2\Big)}_{\eqref{eq:intro-reg-wasserstein}}
+\underbracket{\lambda\|w_0^*\|_2^2}_{\text{constant}}\\[1ex]
&\quad=\underbracket{\Big(\sum_{s,t}\pi_{st}c(s,t)+\lambda_0\|w\|_2\Big)}_{\eqref{eq:unsquared-objective}}
+\underbracket{\lambda\big(\|w\|_2-\|w_0^*\|_2\big)^2}_{\geq0}.
\end{aligned}
\]
The unsquared objective is at least $B$, and the remaining square is nonnegative.
At $(\pi_0^*,w_0^*)$, these terms equal $B$ and zero, respectively.
Since the added constant does not depend on $(\pi,w)$, this pair also minimizes the squared objective.
Every squared minimizer attains the same shifted value $B$, so it must also attain the unsquared minimum $B$.
Since the transport cost is nonnegative,
\[
\RTW_\lambda(S_k)
=\sum_{s,t}\pi_{0,st}^*c(s,t)+\frac{\lambda_0}{2}\|w_0^*\|_2
\quad\Longrightarrow\quad
\RTW_\lambda(S_k)\leq B\leq 2\RTW_\lambda(S_k).
\]

Let $(\pi^*,w^*)$ and $\hat h$ be as in the statement, and let $S_k=S\cup A_k$.
The cost condition~\eqref{eq: def of general dissimilarity} and the transport marginal constraints give $\operatorname{disc}_{\mathcal H,\ell}(\nu_{w^*},\mu)\leq\sum_{s,t}\pi_{st}^*c(s,t)$.
Theorem~\ref{thm:conditional-risk new}, applied with failure probability $\delta/2$, and the triangle inequality for discrepancy therefore give, with conditional probability at least $1-\delta/2$,
\[
\begin{aligned}
L_Q(\hat h)-\inf_{h\in\mathcal H}L_Q(h)
&\leq\sum_{s,t}\pi_{st}^*c(s,t)+\lambda_0\|w^*\|_2
+\operatorname{disc}_{\mathcal H,\ell}(\mu,\mathcal Q_x)\\
&=B+\operatorname{disc}_{\mathcal H,\ell}(\mu,\mathcal Q_x)\\
&\leq2\RTW_\lambda(S_k)+\operatorname{disc}_{\mathcal H,\ell}(\mu,\mathcal Q_x).
\end{aligned}
\]

It remains to bound the discrepancy between the target pool and population.
By definition, $\operatorname{disc}_{\mathcal H,\ell}(\mu,\mathcal Q_x)\leq2\sup_h|L_\mu(h)-L_Q(h)|$.
Introduce auxiliary labels $\widetilde Y_j\sim\mathcal Q_{Y\mid X=t_j}$ independently conditional on $T$, and independent Rademacher signs $\sigma_j$.
Symmetrization over the i.i.d. target features, followed by Jensen's inequality over the auxiliary labels, yields
\[
\begin{aligned}
\mathbb E_T\sup_h|L_\mu(h)-L_Q(h)|
&\leq2\mathbb E_{T,\sigma}\sup_h\left|\frac1N\sum_{j=1}^N\sigma_j g_h(t_j)\right|\\
&\leq2\mathbb E_{T,\widetilde Y,\sigma}\sup_h\left|\frac1N\sum_{j=1}^N\sigma_j\ell(h(t_j),\widetilde Y_j)\right|\\
&\leq\frac{2C_2}{\sqrt N}\left(\sqrt{\operatorname{Pdim}(\mathcal F_\ell)}+1\right),
\end{aligned}
\]
where the last step is the weighted complexity bound from the proof of Theorem~\ref{thm:conditional-risk new}, with uniform weights $1/N$.
Changing one target feature changes $\sup_h|L_\mu(h)-L_Q(h)|$ by at most $1/N$.
McDiarmid's inequality thus gives, for a universal constant $C'$, with probability at least $1-\delta/2$ over $T$,
\[
\operatorname{disc}_{\mathcal H,\ell}(\mu,\mathcal Q_x)
\leq\frac{C'}{\sqrt N}\left(\sqrt{\operatorname{Pdim}(\mathcal F_\ell)}+\sqrt{\log\frac4\delta}\right).
\]
The label bound holds conditionally for every feature realization. Integrating this bound and taking a union bound with the feature event proves the stated probability $1-\delta$; the two events need not be independent.
\end{proof}

\section{Proofs from Section \ref{sec:algorithm}}\label{sec:appendix-algorithm}
In this section we prove the results in Section \ref{sec:algorithm} i.e. Theorem \ref{thm:algorithmic_guarantee}.
Theorem \ref{thm:algorithmic_guarantee} follows from the fact that $\RTW_\lambda$ is a \emph{supermodular} function.   

\begin{theorem}\label{thm:supermodularity_rtw}
    For every $S \subseteq S'$, $x \not\in S'$, we have 
    \begin{equation*}
        \RTW_\lambda(S) - \RTW_\lambda(S \cup \{x\}) \ge \RTW_\lambda(S') - \RTW_\lambda(S' \cup \{x\}).
    \end{equation*}
\end{theorem}

\subsection{Supermodularity Proof}

We first prove Theorem \ref{thm:supermodularity_rtw}. We in fact show that a more general function is supermodular. 
Let $T \subseteq \mathcal{X}$ be the target set of points, and $q \in \R^T_{\ge0}$ be a vector over $T$ with nonnegative entries. For any finite nonempty set of points $S$, define the function $f_S: \R^T_{\ge0} \to \R$ as
\begin{align*}
    f_S(q) = &\min_{\pi_{st} \ge 0, w_s} \sum_{s \in S}\sum_{t \in T} \pi_{st}c(s,t) + \lambda\sum_{s \in S} w_s^2 \\
    &\text{subject to } \sum_{s \in S} \pi_{st} = q_t \forall t \in T, \\
    &\sum_{t \in T}\pi_{st} = w_s, \forall s \in S
\end{align*} where $c(s,t)$ is the cost between points $s \in S$ and $t \in T$ as defined in equation (\ref{eq:intro-reg-wasserstein}), and $\lambda > 0$. We then have that $\RTW_\lambda(S) = f_S(\mu)$ where $\mu \in \R^T_{>0}$ is such that $\mu_t = \frac{1}{\mid T\mid}$.

We prove that $f_S(q)$ is supermodular for any value of $q$. 

\begin{theorem}\label{theorem:supermodularity}
    For all $S \subset S'$ and point $x \not\in S'$,
    \begin{equation*}
        f_{S}(q) - f_{S \cup\{x\}}(q) \ge f_{S'}(q) - f_{S' \cup\{x\}}(q)
    \end{equation*} for all $q \in \R^T_{>0}$.
\end{theorem}

We will use the following lemmas. 

\begin{lemma}\label{lemma:flow_on_x}
    For a set $S$ and a point $x \not\in S$, 
    \begin{align*}
        f_{S \cup \{x\}}(q) = \min_{0 \le \delta \le q} f_{S}(q-\delta) + \sum_{t\in T} c(x,t)\delta_t+ \lambda \inparen{\sum_{t\in T} \delta_t}^2
    \end{align*} for all $q \in \R^T_{\ge0}$.
\end{lemma}

\begin{lemma}\label{lemma:derivative_result}
    The function $f_S$ is differentiable for all $q > 0$. Furthermore, 
    for sets $S,S'$ such that $S \subset S'$,
    \begin{align*}
        \frac{\partial f_S}{\partial q_t}(q) \ge \frac{\partial f_{S'}}{\partial q_t}(q).
    \end{align*} for all $t\in T$, wherever $f_S$ is differentiable. 
\end{lemma}

We first prove Theorem \ref{theorem:supermodularity} assuming Lemma \ref{lemma:flow_on_x} and Lemma \ref{lemma:derivative_result}.

\begin{proof}[Proof of Theorem \ref{theorem:supermodularity}] Using Lemma \ref{lemma:flow_on_x}, we can write 
\begin{align*}
    f_{S}(q) - f_{S \cup \{x\}}(q) &= \max_{0 \le \delta \le q} f_S(q) - f_S(q-\delta) - \sum_{t \in T} \delta_t c(x,t) - \lambda \left(\sum_{t \in T}\delta_t\right)^2 \\
    &= \max_{0 \le \delta \le q} \sum_{t \in T} \int_{0}^1 \delta_t \frac{\partial f_S}{\partial q_t}(q-\tau\delta){\diff}\tau - \sum_{t \in T} \delta_t c(x,t) - \lambda \left(\sum_{t \in T}\delta_t\right)^2 \\
    &= \max_{0 \le \delta \le q} \Delta_S(q,\delta)
\end{align*}

where $\Delta_S(q,\delta) = \sum_{t \in T} \int_{0}^1 \delta_t \frac{\partial f_S}{\partial q_t}(q-\tau\delta)\diff\tau - \sum_{t \in T} \delta_t c(x,t) - \lambda \left(\sum_{j \in T}\delta_j\right)^2$.
If $q-\delta$ has zero coordinates, the integral is understood as the limit as its upper endpoint tends to $1$; continuity of $f_S$ gives the same identity.

By Lemma \ref{lemma:derivative_result}, for any $\delta$, we have that
\begin{align*}
    &\sum_{t \in T} \int_{0}^1 \delta_t \frac{\partial f_S}{\partial q_t}(q-\tau\delta)\diff\tau \ge \sum_{t \in T} \int_{0}^1 \delta_t \frac{\partial f_{S'}}{\partial q_t}(q-\tau\delta)\diff\tau \\
    \implies & \Delta_S(q,\delta) \ge \Delta_{S'}(q,\delta) \\
    \implies & \max_{0 \le \delta \le q} \Delta_S(q,\delta) \ge \max_{0 \le \delta \le q} \Delta_{S'}(q,\delta) \\
    \implies & f_S(q) - f_{S \cup \{x\}}(q) \ge f_{S'}(q) - f_{S' \cup \{x\}}(q). 
\end{align*}
\end{proof}

We now give the proof of the helper lemmas. 

\begin{proof}[Proof of Lemma \ref{lemma:flow_on_x}]
    We can write the optimization problem for $S \cup \{x\}$ as 

    \begin{align*}
        f_{S \cup \{x\}}(q) &= \min_{\pi_{st},\pi_{xt}\ge0,\,w,w_x} \sum_{s \in S}\sum_{t \in T} \pi_{st}c(s,t) + \lambda \sum_{s \in S}    w_s^2 +
        \sum_{t \in T} \pi_{xt} c(x,t) + \lambda w_x^2 \\
        & \text{ s.t. } \forall t\in T, \sum_{s\in S}\pi_{st} + \pi_{xt} = q_t, \forall s \in S, \sum_{t \in T} \pi_{st} = w_s,
        \text{ and } \sum_{t \in T} \pi_{xt} = w_x.
    \end{align*}
        Introduce an auxiliary variable $\delta \in \R^T$ satisfying $\delta_t = q_t - \sum_{s \in S} \pi_{st}$ for all $t \in T$. Then, $\pi_{xt}=\delta_t$ for all $t \in T$, and we can write an equivalent optimization problem:
        \begin{align*}
        f_{S \cup \{x\}}(q) &= \min_{0 \le \delta \le q} \min_{\pi_{st}\ge0,\,w_s} \sum_{s \in S} \sum_{t \in T} \pi_{st}c(s,t) + \lambda \sum_{s \in S}    w_s^2 +
        \sum_{t \in T} \delta_{t} c(x,t) +  \lambda \inparen{\sum_{t \in T} \delta_t}^2 \\
        & \text{ s.t. } \forall t\in T, \sum_{s\in S}\pi_{st}= q_t - \delta_t, \forall s \in S, \sum_{t \in T} \pi_{st} = w_s.
        \end{align*}
        Finally, the internal optimization problem over $\pi, w$ can be written in the form required. 
\end{proof}

\begin{proof}[Proof of Lemma \ref{lemma:derivative_result}]
   The objective function is strictly convex in $w$, and this ensures that the optimal weights are unique. Thus, Danskin's theorem \citep{bertsekas2009convex} implies that $f_S$ is differentiable. 

    Further, we can write the derivative using the KKT conditions as 
    \begin{align*}
        \frac{\partial f_S}{\partial q_t}(q) = \min_{\ell \in S} c(\ell,t)+2 \lambda \cdot  w_\ell^*(S)
    \end{align*} where $w^*_\ell(S)$ is the optimal assignment for the $w$ variables. From the KKT conditions, we also have, for each $t \in T$,  
    \begin{align}\label{equation:complementary_slackness}
        \pi_{st}^*(S) > 0 \implies c(s,t) + 2\lambda \cdot w_s^*(S) = \min_{\ell \in S} c(\ell,t)+2 \lambda \cdot  w_\ell^*(S)
    \end{align}
    For $S' \supset S$, we get 
    \begin{align}\label{equation:derivative_of_Sprime}
        \frac{\partial f_{S'}}{\partial q_t}(q) &= \min_{\ell \in S'} c(\ell,t) + 2 \lambda \cdot w_\ell^*(S') \\
        &\le \min_{\ell \in S} c(\ell,t) + 2 \lambda \cdot w_\ell^*(S').
    \end{align}
    We will now show that $w^*_{\ell}(S') \le w^*_\ell(S)$, for all $\ell \in S$, from which the result follows. 

    Let $A = \{\ell \in S: w^*_{\ell}(S') > w^*_{\ell}(S)\}$, and assume to the contrary that $A$ is non-empty. Then, since $\sum_{s \in A}w_s^*(S') > \sum_{s \in A} w_s^*(S)$, there exists $t \in T, \ell \in A$ and $k \in S \setminus A$ such that $\pi^*_{\ell t}(S') > \pi_{\ell t}^*(S)$ and $\pi_{kt}^*(S') < \pi_{kt}^*(S)$. This implies that $\pi_{\ell t}^*(S') > 0$ and $\pi_{kt}^*(S) > 0$, and equation (\ref{equation:complementary_slackness}) implies that 
    \begin{align*}
        c(\ell,t) + 2\lambda\cdot w_\ell^*(S') \le c(k,t) + 2\lambda\cdot w_k^*(S') &\text{ and } c(\ell,t) + 2\lambda\cdot w_\ell^*(S) \ge c(k,t) + 2\lambda\cdot w_k^*(S).
    \end{align*}
    Since $\ell \in A$, $w_\ell^*(S') > w_{\ell}^*(S)$. Combined with the above, we have $w_k^*(S') > w_{k}^*(S)$.
    This is a contradiction, since $k \not\in A$, we must have $w_k^*(S') \le w_{k}^*(S)$. 

    Hence, we must have that $w^*_{\ell}(S') \le w^*_\ell(S)$, for all $\ell \in S$. Using equation (\ref{equation:derivative_of_Sprime}), we get 
    \begin{align*}
    \frac{\partial f_{S'}}{\partial q_t}(q) &\le \min_{\ell \in S} c(\ell,t) + 2 \lambda \cdot w_\ell^*(S') \\
    &\le \min_{\ell \in S} c(\ell,t) + 2\lambda \cdot w_\ell^*(S) \\
    &= \frac{\partial f_{S}}{\partial q_t}(q).
    \end{align*}
\end{proof}

\subsection{Greedy approximation guarantee}\label{app:greedy-guarantee}
\begin{proof}[Proof of Theorem~\ref{thm:algorithmic_guarantee}]
Consider the function  $g(A):= \RTW_\lambda(S)-\RTW_\lambda(S\cup A)$. By Theorem~\ref{thm:supermodularity_rtw}, $g$ is monotone and submodular. The first claim therefore follows directly from the cardinality-constrained greedy guarantee of~\citet{nemhauser1978analysis}.

For the second claim, consider greedily adding another queried point to $S\cup\hat A_i$ on each step $i$, and suppose the current objective is still larger than the value given by $A_k^*$, that is,
\[ \RTW_\lambda(S\cup \hat A_i) \geq \RTW_\lambda (S \cup A_k^*). \]

By the greedy property and diminishing returns from adding each point (supermodularity), the next greedy step reduces this cost difference by at least $1/k$. 
Formally, we have
\[
\RTW_\lambda(S\cup\hat A_{i+1})-\RTW_\lambda(S\cup A_k^*)
\leq \left(1-\frac1k\right)
\inparen{\RTW_\lambda(S\cup\hat A_i)-\RTW_\lambda(S\cup A_k^*)}.
\]
If the current cost is already at most $\RTW_\lambda(S\cup A_k^*)$, the desired guarantee holds by monotonicity. Otherwise, iterating from $\hat A_0=\varnothing$ gives
\[
\RTW_\lambda(S\cup\hat A_{k'})
\leq \RTW_\lambda(S\cup A_k^*)
+e^{-k'/k}\inparen{\RTW_\lambda(S)-\RTW_\lambda(S\cup A_k^*)}. 
\]
If $k'=|T|$, all target points are selected, so monotonicity gives $\RTW_\lambda(S\cup\hat A_{k'})\leq\RTW_\lambda(S\cup A_k^*)\leq\eta\RTW_\lambda(S)$.
Otherwise, $k'=\lceil k\log(1/(\epsilon\eta))\rceil$, and the last term is at most $\epsilon\eta\RTW_\lambda(S)$.
Combining this with $\RTW_\lambda(S\cup A_k^*)\leq\eta\RTW_\lambda(S)$ proves the second claim, including when the optimal cost is zero.
\end{proof}

\section{Additional Experimental Details} 
\label{sec:appendix-experiments}
In this section we present the full experimental results, and mention some additional implementation details. 

\paragraph{Additional Implementation Details.} To speed up our algorithm, we use several computational techniques. Since we solve a sequence of optimization problems with an increasing number of variables, we warm-start each problem using the solution to the preceding problem. For example, when computing $\RTW_\lambda(S \cup \{x\})$, we initialize the variables corresponding to $S$ using the solution obtained while computing $\RTW_\lambda(S)$. Additionally, at each iteration, we first screen the candidate points by solving a one-dimensional optimization problem for each candidate to obtain a dual lower bound. We then solve the full optimization problem only for the candidates that cannot be ruled out by these lower bounds. Finally, we solve multiple optimization problems in parallel batches on a GPU. 

\begin{table*}[t]
\centering
\begingroup
\renewcommand{\arraystretch}{1.2}
\resizebox{\textwidth}{!}{%
\begin{tabular}{ccccccccc}
\hline
Task & $K$ & RTW ($\lambda=10$) & K-medoids+W & MaxHerd & K-medoids & BADGE & CLUE & Random \\
\hline
Clipart $\to$ Product & 50 & $67.66 \pm 1.05$ & {\bfseries\boldmath $69.07 \pm 1.17$} & \underline{$68.41 \pm 0.88$} & $68.31 \pm 2.40$ & $58.89 \pm 2.12$ & $50.38 \pm 3.84$ & $59.24 \pm 1.80$ \\
 & 100 & $72.85 \pm 1.35$ & {\bfseries\boldmath $74.66 \pm 1.03$} & \underline{$73.90 \pm 1.86$} & $73.50 \pm 1.59$ & $68.46 \pm 1.93$ & $58.49 \pm 3.32$ & $66.20 \pm 1.98$ \\
\hline
Clipart $\to$ Real World & 50 & $67.89 \pm 2.83$ & \underline{$68.05 \pm 2.17$} & {\bfseries\boldmath $68.13 \pm 2.50$} & $67.33 \pm 2.70$ & $61.99 \pm 3.42$ & $53.94 \pm 2.94$ & $61.35 \pm 3.32$ \\
 & 100 & $71.39 \pm 1.55$ & {\bfseries\boldmath $72.75 \pm 1.82$} & $71.55 \pm 2.04$ & \underline{$71.79 \pm 2.21$} & $68.84 \pm 2.60$ & $61.35 \pm 2.25$ & $66.45 \pm 2.49$ \\
\hline
Product $\to$ Clipart & 50 & {\bfseries\boldmath $39.12 \pm 3.18$} & \underline{$39.02 \pm 3.17$} & $37.84 \pm 1.06$ & $37.65 \pm 3.07$ & $36.18 \pm 3.13$ & $33.33 \pm 2.08$ & $36.27 \pm 3.47$ \\
 & 100 & \underline{$43.53 \pm 2.41$} & {\bfseries\boldmath $44.02 \pm 2.78$} & $42.16 \pm 2.38$ & $41.76 \pm 2.18$ & $40.88 \pm 4.62$ & $38.73 \pm 2.59$ & $39.80 \pm 3.49$ \\
\hline
Product $\to$ Real World & 50 & {\bfseries\boldmath $75.78 \pm 2.15$} & $72.91 \pm 2.44$ & $73.31 \pm 2.83$ & \underline{$73.39 \pm 3.23$} & $68.84 \pm 3.24$ & $66.61 \pm 2.72$ & $68.69 \pm 2.95$ \\
 & 100 & {\bfseries\boldmath $76.73 \pm 2.61$} & $74.98 \pm 2.46$ & $75.46 \pm 2.73$ & \underline{$75.86 \pm 3.45$} & $72.35 \pm 2.65$ & $70.52 \pm 5.00$ & $71.16 \pm 1.63$ \\
\hline
Real World $\to$ Clipart & 50 & {\bfseries\boldmath $40.88 \pm 2.61$} & $40.59 \pm 2.15$ & \underline{$40.88 \pm 1.41$} & $39.22 \pm 2.38$ & $37.94 \pm 1.98$ & $35.69 \pm 2.58$ & $37.84 \pm 4.68$ \\
 & 100 & {\bfseries\boldmath $45.29 \pm 2.47$} & $44.02 \pm 2.89$ & $44.12 \pm 3.16$ & $43.92 \pm 2.85$ & \underline{$44.61 \pm 2.19$} & $40.20 \pm 2.94$ & $40.00 \pm 3.61$ \\
\hline
Real World $\to$ Product & 50 & {\bfseries\boldmath $75.87 \pm 1.94$} & $71.13 \pm 2.20$ & $72.19 \pm 2.38$ & \underline{$72.19 \pm 1.78$} & $68.61 \pm 2.24$ & $62.92 \pm 4.01$ & $67.86 \pm 2.80$ \\
 & 100 & {\bfseries\boldmath $77.78 \pm 1.55$} & $76.78 \pm 2.10$ & \underline{$77.13 \pm 0.57$} & $76.73 \pm 1.60$ & $72.70 \pm 1.11$ & $68.72 \pm 2.58$ & $71.69 \pm 2.72$ \\
\hline
\end{tabular}%
}
\endgroup
\caption{Held-out target accuracies for active domain adaptation methods on all OfficeHome-RSUT tasks, at two query budgets $K = 50$ and $K = 100$. The values displayed are mean $\pm$ standard deviation across five random 80/20 acquisition pool and evaluation splits. The best values in each row are indicated in bold, and the second best values are underlined. }
\label{tab:officehome-rsut-active}
\end{table*}

\paragraph{Baselines. } As described in Section \ref{sec:experiments}, our baselines can broadly be divided into importance-weighting baselines and active domain adaptation baselines. We consider three importance-weighting baselines. \textbf{Kernel Mean Matching (KMM)} (\citet{huang2007selectionBias}) estimates the importance weights by minimizing the MMD between the (reweighted) source and the target dataset. \textbf{ KLIEP} (\citet{sugiyama2007kliep}) computes importance weights by minimizing the KL divergence between the train and test distributions.\textbf{ Importance Weighting Network (IWN)} (\citet{demathelin2022iwn}) uses a neural network to learn the importance weights. 

We considered five active domain adaptation methods. \citet{deMathelin2022discrepancy} use \textbf{K-medoids} to select the points to query. They also introduce \textbf{K-medoids+W}, a weighted version of $k$-medoids that uses the source labels to assign larger weights to points closer to the margin. Note that \textbf{K-medoids}, with nearest-neighbor reweighting, corresponds to our method with $\lambda = 0$. 
\textbf{MaxHerd} (\citet{bae2024maxherd}) greedily queries points to maximize the kernel-based coverage of the target samples.
\textbf{BADGE} (\citet{ash2020badge}) uses the gradient embeddings to pick points that are uncertain and diverse. \textbf{CLUE} (\citet{prabhu2021clue}) uses uncertainty-weighted clustering to pick target points that are both uncertain and diverse. 
In addition to these, we also consider \textbf{Random}, which queries points uniformly at random from the target. 

\begin{table}[t]
\centering
\begingroup
\renewcommand{\arraystretch}{1.2}
\resizebox{\columnwidth}{!}{%
\begin{tabular}{ccccc}
\hline
Task & RTW ($\lambda=10$) & KMM & KLIEP & IWN \\
\hline
Clipart $\to$ Product & {\bfseries\boldmath $49.02 \pm 2.08$} & $30.63 \pm 1.96$ & \underline{$46.55 \pm 1.43$} & $43.17 \pm 2.24$ \\
Clipart $\to$ Real World & {\bfseries\boldmath $53.15 \pm 0.96$} & $25.82 \pm 1.81$ & \underline{$50.36 \pm 1.11$} & $45.90 \pm 2.81$ \\
Product $\to$ Clipart & \underline{$31.57 \pm 2.34$} & $13.92 \pm 2.61$ & {\bfseries\boldmath $32.06 \pm 3.61$} & $26.86 \pm 2.26$ \\
Product $\to$ Real World & {\bfseries\boldmath $69.24 \pm 2.13$} & $56.02 \pm 3.81$ & $61.43 \pm 1.45$ & \underline{$64.38 \pm 2.74$} \\
Real World $\to$ Clipart & \underline{$32.16 \pm 2.99$} & $15.20 \pm 1.47$ & {\bfseries\boldmath $33.82 \pm 2.86$} & $27.65 \pm 2.04$ \\
Real World $\to$ Product & {\bfseries\boldmath $68.16 \pm 1.00$} & $57.73 \pm 1.95$ & \underline{$64.03 \pm 1.03$} & $63.73 \pm 1.69$ \\
\hline
\end{tabular}%
}
\endgroup
\caption{Held-out target accuracies for importance-weighting baselines on OfficeHome-RSUT tasks. The values displayed are mean $\pm$ standard deviation across five random 80/20 acquisition pool and evaluation splits. Each method trains on the labeled source samples using its own learned reweighting. The best values in each row are indicated in bold, and the second best values are underlined.}
\label{tab:officehome-rsut-importance}
\end{table}

\paragraph{OfficeHome-RSUT. } Table (\ref{tab:officehome-rsut-active}) shows accuracy of all active domain adaptation methods on all OfficeHome-RSUT tasks, evaluated at two query budgets. Our method outperforms or is competitive against other active domain adaptation baselines. Table (\ref{tab:officehome-rsut-importance}) shows accuracy of all the importance weighting methods, and compares it to the performance of our reweighting at no additional queries. RTW outperforms or is competitive with all other importance-weighting methods. 

\begin{figure}[t]
    \centering
    \includegraphics[width=\linewidth]{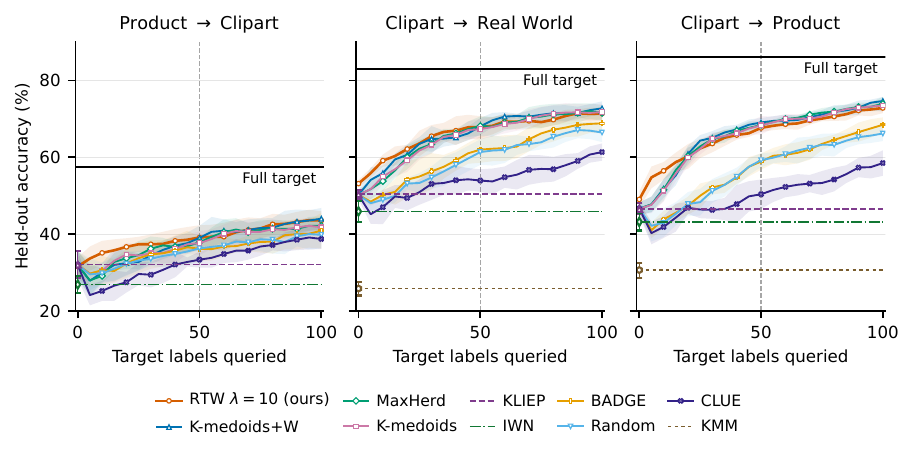}
    \caption{
    Results for three additional tasks on OfficeHome-RSUT. The full target line denotes the accuracy of the classifier trained on the target query pool. Dashed lines represent methods that do not query for additional labels.
    }\label{fig:officehome_rsut_appendix}
\end{figure}

Figure (\ref{fig:officehome_rsut_appendix}) shows the accuracy of different methods on three additional tasks. Our RTW algorithm is competitive with other baselines at $K = 100$, and outperforms baselines for smaller query budgets.

\begin{figure}[t]
    \centering
    \includegraphics[width=\linewidth]{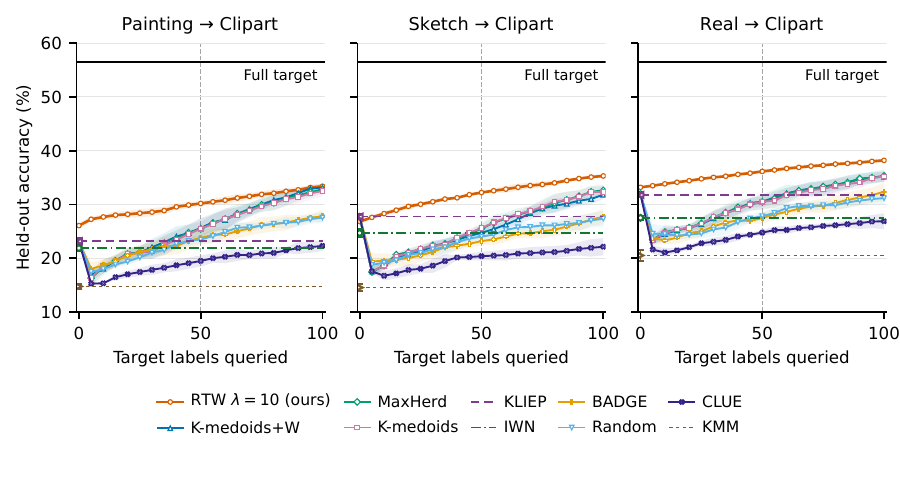}
    \caption{Held-out accuracy on the three mini-DomainNet tasks with Clipart as the target domain (RTW with $\lambda=10$). Curves and shaded bands show means and one sample standard deviation across five splits, and panels share the same y-axis range. Full target denotes the accuracy of a classifier trained on the entire query pool.}
    \label{fig:minidomainnet-to-clipart}
\end{figure}

\begin{figure}[t]
    \centering
    \includegraphics[width=\linewidth]{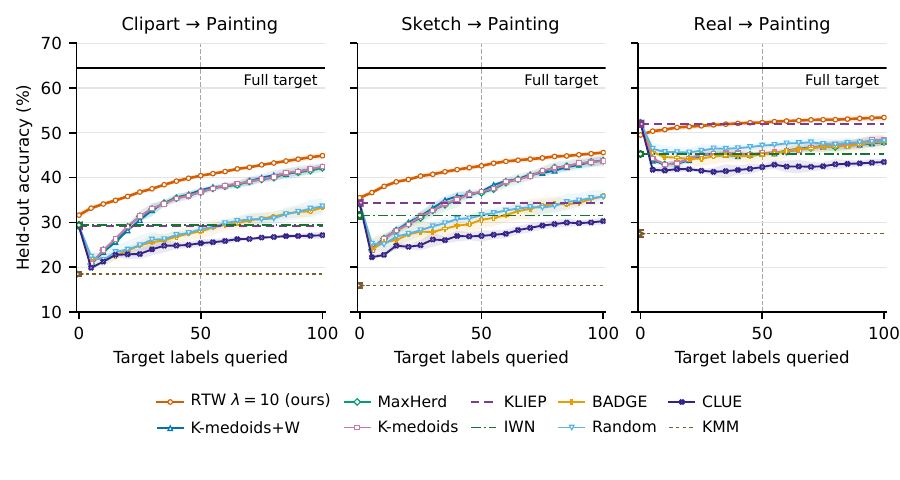}
    \caption{Held-out accuracy on the three mini-DomainNet tasks with Painting as the target domain, following the conventions of Figure~\ref{fig:minidomainnet-to-clipart}.}
    \label{fig:minidomainnet-to-painting}
\end{figure}

\begin{figure}[t]
    \centering
    \includegraphics[width=\linewidth]{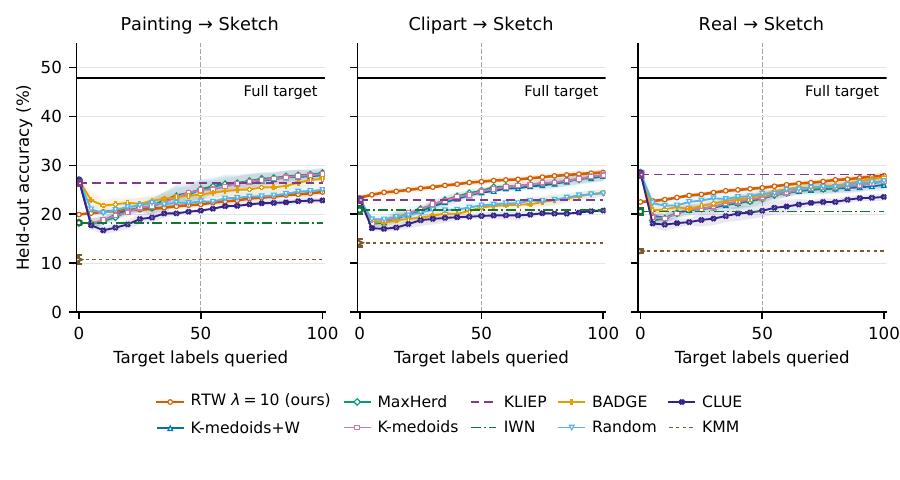}
    \caption{Held-out accuracy on the three mini-DomainNet tasks with Sketch as the target domain, following the conventions of Figure~\ref{fig:minidomainnet-to-clipart}.}
    \label{fig:minidomainnet-to-sketch}
\end{figure}

\paragraph{mini-DomainNet.} Table~(\ref{tab:minidomainnet-active}) reports query-method accuracies on all 12 tasks at $K=50$ and $K=100$, and Table~(\ref{tab:minidomainnet-importance}) compares source reweighting without additional queries on the same tasks.
Figures~\ref{fig:minidomainnet-to-clipart}--\ref{fig:minidomainnet-to-sketch} show the nine tasks whose target is not Real. Across the 24 task--budget pairs in Table~(\ref{tab:minidomainnet-active}), RTW has the highest accuracy among query methods in 2. The only exceptions are Sketch$\to$Real at $K=100$ and Painting$\to$Sketch at both budgets. On Real$\to$Sketch, RTW also remains below the zero-query KLIEP baseline.

\begin{table*}[t]
\centering
\begingroup
\renewcommand{\arraystretch}{1.2}
\resizebox{\textwidth}{!}{%
\begin{tabular}{ccccccccc}
\hline
Task & $K$ & RTW ($\lambda=10$) & K-medoids+W & MaxHerd & K-medoids & BADGE & CLUE & Random \\
\hline
Clipart $\to$ Real & 50 & {\bfseries\boldmath $62.26 \pm 0.52$} & \underline{$55.32 \pm 0.66$} & $53.76 \pm 0.55$ & $53.94 \pm 0.51$ & $42.30 \pm 2.31$ & $35.67 \pm 0.86$ & $41.54 \pm 0.97$ \\
 & 100 & {\bfseries\boldmath $66.19 \pm 0.28$} & \underline{$63.95 \pm 0.52$} & $63.86 \pm 0.52$ & $63.92 \pm 0.65$ & $49.48 \pm 1.53$ & $40.72 \pm 1.63$ & $47.32 \pm 0.84$ \\
\hline
Painting $\to$ Real & 50 & {\bfseries\boldmath $71.62 \pm 0.27$} & \underline{$62.29 \pm 0.37$} & $61.89 \pm 0.68$ & $61.45 \pm 0.64$ & $56.27 \pm 1.58$ & $47.52 \pm 0.95$ & $53.01 \pm 1.90$ \\
 & 100 & {\bfseries\boldmath $72.87 \pm 0.29$} & \underline{$68.83 \pm 0.43$} & $68.47 \pm 0.75$ & $68.36 \pm 0.54$ & $59.69 \pm 1.92$ & $50.72 \pm 1.22$ & $56.94 \pm 2.09$ \\
\hline
Sketch $\to$ Real & 50 & {\bfseries\boldmath $59.00 \pm 0.48$} & \underline{$51.57 \pm 0.34$} & $51.16 \pm 0.81$ & $51.07 \pm 0.64$ & $40.78 \pm 1.72$ & $31.98 \pm 0.64$ & $40.15 \pm 0.77$ \\
 & 100 & $63.05 \pm 0.41$ & {\bfseries\boldmath $64.62 \pm 0.48$} & $64.38 \pm 0.63$ & \underline{$64.45 \pm 0.62$} & $48.00 \pm 1.98$ & $37.90 \pm 1.55$ & $46.73 \pm 0.67$ \\
\hline
Painting $\to$ Clipart & 50 & {\bfseries\boldmath $30.19 \pm 0.51$} & \underline{$25.62 \pm 1.86$} & $25.46 \pm 1.30$ & $25.48 \pm 1.87$ & $23.78 \pm 1.25$ & $19.50 \pm 1.05$ & $23.49 \pm 0.76$ \\
 & 100 & {\bfseries\boldmath $33.39 \pm 0.41$} & \underline{$33.13 \pm 0.72$} & $32.79 \pm 0.77$ & $32.51 \pm 1.00$ & $27.81 \pm 1.11$ & $22.30 \pm 0.93$ & $27.57 \pm 1.04$ \\
\hline
Sketch $\to$ Clipart & 50 & {\bfseries\boldmath $32.23 \pm 0.23$} & $24.72 \pm 0.84$ & $25.50 \pm 1.06$ & \underline{$25.59 \pm 0.67$} & $23.19 \pm 0.80$ & $20.39 \pm 0.80$ & $24.02 \pm 1.08$ \\
 & 100 & {\bfseries\boldmath $35.31 \pm 0.41$} & $31.82 \pm 0.85$ & \underline{$32.62 \pm 1.01$} & $32.35 \pm 1.06$ & $27.72 \pm 1.42$ & $22.14 \pm 1.59$ & $27.35 \pm 1.45$ \\
\hline
Real $\to$ Clipart & 50 & {\bfseries\boldmath $36.15 \pm 0.37$} & \underline{$30.51 \pm 1.00$} & $30.47 \pm 1.45$ & \underline{$30.51 \pm 1.32$} & $27.47 \pm 1.02$ & $24.77 \pm 0.50$ & $27.82 \pm 0.71$ \\
 & 100 & {\bfseries\boldmath $38.21 \pm 0.36$} & $35.18 \pm 0.96$ & \underline{$35.34 \pm 1.16$} & $35.23 \pm 0.90$ & $32.35 \pm 1.72$ & $26.88 \pm 1.49$ & $31.19 \pm 1.07$ \\
\hline
Clipart $\to$ Painting & 50 & {\bfseries\boldmath $40.44 \pm 0.39$} & \underline{$37.21 \pm 0.66$} & $36.74 \pm 0.69$ & $36.81 \pm 0.58$ & $27.96 \pm 1.62$ & $25.38 \pm 0.86$ & $28.48 \pm 0.65$ \\
 & 100 & {\bfseries\boldmath $44.89 \pm 0.70$} & $42.31 \pm 0.81$ & $42.20 \pm 0.87$ & \underline{$42.46 \pm 0.76$} & $33.36 \pm 1.01$ & $27.14 \pm 0.78$ & $33.59 \pm 1.81$ \\
\hline
Sketch $\to$ Painting & 50 & {\bfseries\boldmath $42.62 \pm 0.59$} & \underline{$36.92 \pm 0.40$} & $36.63 \pm 0.57$ & $36.87 \pm 0.37$ & $30.58 \pm 1.90$ & $26.98 \pm 1.24$ & $31.63 \pm 1.21$ \\
 & 100 & {\bfseries\boldmath $45.59 \pm 0.48$} & \underline{$43.85 \pm 0.95$} & $43.77 \pm 1.29$ & $43.75 \pm 1.08$ & $35.84 \pm 0.61$ & $30.30 \pm 0.83$ & $35.74 \pm 1.53$ \\
\hline
Real $\to$ Painting & 50 & {\bfseries\boldmath $52.27 \pm 0.41$} & $45.34 \pm 0.38$ & $45.40 \pm 0.33$ & $45.50 \pm 0.21$ & $45.13 \pm 1.63$ & $42.35 \pm 0.70$ & \underline{$47.16 \pm 0.63$} \\
 & 100 & {\bfseries\boldmath $53.42 \pm 0.55$} & $47.93 \pm 0.71$ & $47.97 \pm 0.80$ & \underline{$48.43 \pm 1.03$} & $47.94 \pm 1.47$ & $43.48 \pm 0.76$ & $48.16 \pm 1.02$ \\
\hline
Clipart $\to$ Sketch & 50 & {\bfseries\boldmath $26.67 \pm 0.64$} & $24.45 \pm 0.53$ & \underline{$24.81 \pm 0.62$} & $24.69 \pm 0.63$ & $21.39 \pm 0.79$ & $19.65 \pm 0.67$ & $21.81 \pm 0.68$ \\
 & 100 & {\bfseries\boldmath $28.54 \pm 0.60$} & $27.84 \pm 0.99$ & $27.95 \pm 1.34$ & \underline{$27.99 \pm 1.08$} & $24.42 \pm 0.68$ & $20.77 \pm 0.52$ & $24.26 \pm 0.68$ \\
\hline
Painting $\to$ Sketch & 50 & $22.02 \pm 0.56$ & \underline{$25.02 \pm 2.13$} & {\bfseries\boldmath $25.14 \pm 1.85$} & $24.96 \pm 1.82$ & $24.01 \pm 1.01$ & $20.75 \pm 0.36$ & $22.65 \pm 0.96$ \\
 & 100 & $24.55 \pm 0.62$ & $27.99 \pm 1.04$ & {\bfseries\boldmath $28.44 \pm 1.07$} & \underline{$28.22 \pm 1.23$} & $27.25 \pm 1.25$ & $22.83 \pm 0.59$ & $24.97 \pm 0.89$ \\
\hline
Real $\to$ Sketch & 50 & {\bfseries\boldmath $25.41 \pm 0.56$} & $23.25 \pm 1.13$ & $23.19 \pm 1.28$ & $23.39 \pm 1.32$ & $23.93 \pm 1.14$ & $20.72 \pm 1.30$ & \underline{$24.25 \pm 0.77$} \\
 & 100 & {\bfseries\boldmath $27.88 \pm 0.59$} & $26.01 \pm 1.21$ & $26.65 \pm 1.40$ & $26.77 \pm 1.52$ & \underline{$27.49 \pm 1.15$} & $23.55 \pm 0.73$ & $26.77 \pm 0.30$ \\
\hline
\end{tabular}%
}
\endgroup
\caption{Held-out target accuracies (\%) for active domain adaptation methods on all 12 mini-DomainNet tasks, grouped by target domain, at query budgets $K=50$ and $K=100$. Values are mean $\pm$ sample standard deviation across five random 80/20 acquisition-pool and evaluation partitions. The best value in each row is bold, and the second best is underlined.}
\label{tab:minidomainnet-active}
\end{table*}

\begin{table}[t]
\centering
\begingroup
\renewcommand{\arraystretch}{1.2}
\resizebox{\columnwidth}{!}{%
\begin{tabular}{ccccc}
\hline
Task & RTW ($\lambda=10$) & KMM & KLIEP & IWN \\
\hline
Clipart $\to$ Real & {\bfseries\boldmath $51.72 \pm 0.14$} & $37.69 \pm 0.88$ & $42.62 \pm 0.18$ & \underline{$44.84 \pm 0.28$} \\
Painting $\to$ Real & {\bfseries\boldmath $68.64 \pm 0.30$} & $44.91 \pm 2.57$ & \underline{$57.62 \pm 0.28$} & $56.76 \pm 0.52$ \\
Sketch $\to$ Real & {\bfseries\boldmath $50.76 \pm 0.15$} & $22.70 \pm 0.61$ & \underline{$43.58 \pm 0.38$} & $40.10 \pm 1.15$ \\
\hline
Painting $\to$ Clipart & {\bfseries\boldmath $26.07 \pm 0.49$} & $14.75 \pm 0.52$ & \underline{$23.16 \pm 0.57$} & $21.89 \pm 0.44$ \\
Sketch $\to$ Clipart & \underline{$26.82 \pm 0.37$} & $14.54 \pm 0.65$ & {\bfseries\boldmath $27.75 \pm 0.46$} & $24.69 \pm 0.76$ \\
Real $\to$ Clipart & {\bfseries\boldmath $33.19 \pm 0.26$} & $20.52 \pm 0.98$ & \underline{$31.77 \pm 0.33$} & $27.52 \pm 0.35$ \\
\hline
Clipart $\to$ Painting & {\bfseries\boldmath $31.62 \pm 0.66$} & $18.47 \pm 0.49$ & $29.26 \pm 0.28$ & \underline{$29.40 \pm 0.54$} \\
Sketch $\to$ Painting & {\bfseries\boldmath $35.51 \pm 0.46$} & $15.92 \pm 0.49$ & \underline{$34.39 \pm 0.47$} & $31.57 \pm 0.82$ \\
Real $\to$ Painting & \underline{$49.52 \pm 0.56$} & $27.49 \pm 0.74$ & {\bfseries\boldmath $51.97 \pm 0.56$} & $45.27 \pm 0.35$ \\
\hline
Clipart $\to$ Sketch & {\bfseries\boldmath $23.38 \pm 0.46$} & $14.16 \pm 0.75$ & \underline{$22.92 \pm 0.60$} & $20.85 \pm 0.78$ \\
Painting $\to$ Sketch & \underline{$19.94 \pm 0.25$} & $10.72 \pm 0.86$ & {\bfseries\boldmath $26.35 \pm 0.59$} & $18.27 \pm 0.42$ \\
Real $\to$ Sketch & \underline{$22.52 \pm 0.51$} & $12.46 \pm 0.46$ & {\bfseries\boldmath $28.11 \pm 0.61$} & $20.57 \pm 0.76$ \\
\hline
\end{tabular}%
}
\endgroup
\caption{Held-out target accuracies (\%) for importance-weighting methods on all 12 mini-DomainNet tasks at $K=0$. Values are mean $\pm$ sample standard deviation across random 80/20 partitions of the target into acquisition pool and held-out evaluation set. Each method trains only on labeled source samples using its learned weights. The best value in each row is bold, and the second best is underlined.}
\label{tab:minidomainnet-importance}
\end{table}

\subsection{Synthetic Experiments}
\label{sec:synthetic_experiments}

We present some synthetic experiments which concretely demonstrate situations where other methods can fail, while our algorithm performs well. In particular, the experiments demonstrate failure points of the $k$-medoids based query strategies. 

\paragraph{Dataset and Task. } Our synthetic experiments are regression tasks. The feature space $\mathcal{X}$ is $\R^{40}$, while the label space $\mathcal{Y}$ is $\mathbb{R}$. 

The source distribution $\mathcal{P}_x$ is an isotropic  Gaussian distribution $\mathcal{N}(-12e_3, I)$, where $e_3$ is the third standard unit vector. The target distribution $\mathcal{Q}_x$ is a mixture of three distributions - with 0.1 probability, we sample $\mathcal{N}(-12e_3, I)$, with 0.5 probability we sample the point $12e_3$, and with probability 0.4, we sample the origin. 

The label distribution is as follows:
\begin{equation*}
    Y | x = \begin{cases}
  \mathcal{N}(-1,1/4) & \text{if } x = 12e_3 \\
  \mathcal{N}(1,1/4) & \text{if } x = 0 \\
  0 & \text{otherwise.}
\end{cases}
\end{equation*}

The source dataset consists of 200 labeled samples, while the target dataset consists of 1000 unlabeled samples. For each method, we train a regressor using ridge regression, on the labeled source samples together with any queried target samples, with the weighting prescribed by the method. 

We run thirty trials of our experiment. We report the mean-squared error on the target dataset as the performance metric, at query budgets $K = 10, 20, 30$. We run the experiment for the active learning baselines. Our RTW method uses $\lambda = 1$. 

\paragraph{Results and Discussion. } Table \ref{tab:synthetic_experiments_result} reports the mean squared error of our experiments. The baselines perform only slightly better than random, whereas our method performs substantially better. 

These synthetic experiments demonstrate a concrete situation where other methods might fail. Qualitatively, $k$-medoids queries the labels at the noisy locations in the target distribution (i.e. the origin and $12e_3$) only once, since choosing additional queries cannot further decrease the $k$-medoids cost. In contrast, RTW chooses to query those locations multiple times, which can improve estimation of the labels at each location.

While we described a specific setting for concreteness, these failure modes for $k$-medoids query selection are natural in settings with very heterogeneous populations with clusters of different radii, e.g. a mixture of Gaussians with different covariances. $k$-medoids picks at most one (or very few) points from the tighter clusters, even if there is variation in the labels. This problem is exacerbated in high dimensions where strong distance concentration will lead to the queried points being almost exclusively from the higher variance cluster. 

\begin{table}[t]
\centering
\scriptsize
\setlength{\tabcolsep}{3pt}
\begin{tabular}{lcrrrrrr}
\hline
$K$ & RTW ($\lambda=1$) & K-medoids & MaxHerd & BADGE-R & CLUE-R & Random \\
\hline
10 & $\mathbf{0.797 \pm 0.015}$ & $1.087 \pm 0.041$ & $1.087 \pm 0.041$ & $1.059 \pm 0.037$ & $\underline{0.951 \pm 0.012}$ & $1.066 \pm 0.026$ \\
20 & $\mathbf{0.782 \pm 0.014}$ & $1.099 \pm 0.040$ & $1.099 \pm 0.040$ & $1.006 \pm 0.021$ & $\underline{0.974 \pm 0.010}$ & $0.991 \pm 0.009$ \\
30 & $\mathbf{0.776 \pm 0.012}$ & $1.100 \pm 0.039$ & $1.100 \pm 0.039$ & $0.996 \pm 0.012$ & $\underline{0.966 \pm 0.007}$ & $0.976 \pm 0.006$ \\
\hline
\end{tabular}
\caption{Mean squared error for synthetic regression experiments for different values of the query budget $K$. Values are mean $\pm$ sample standard deviation across thirty random trials of the experiment. The best value in each row is bold, and the second best is underlined.}
\label{tab:synthetic_experiments_result}
\end{table}

\end{document}